\documentclass[a4paper,12pt]{report}
\usepackage[utf8x]{inputenc}

\usepackage{amsmath} 
\usepackage{amssymb} 
\usepackage{graphicx} 
\usepackage{subcaption}
\usepackage{verbatim}
\usepackage{mathtools}
\usepackage{subcaption}
\usepackage[titlenumbered,ruled,linesnumbered]{algorithm2e}
\usepackage{todonotes}
\usepackage{booktabs}

\usepackage[UKenglish]{babel}

\usepackage{hyperref}
\hypersetup{
    citecolor=black,
    filecolor=black,
    linkcolor=green,
    urlcolor=black
}
\usepackage{flexisym}
\usepackage{listings}
\usepackage{color}
\usepackage{float}
\usepackage{wrapfig}

\definecolor{dkgreen}{rgb}{0,0.6,0}
\definecolor{gray}{rgb}{0.5,0.5,0.5}
\definecolor{mauve}{rgb}{0.58,0,0.82}

\begin{document}

\begin{titlepage}

\center 


\vspace{-1cm}
\includegraphics[width=4cm]{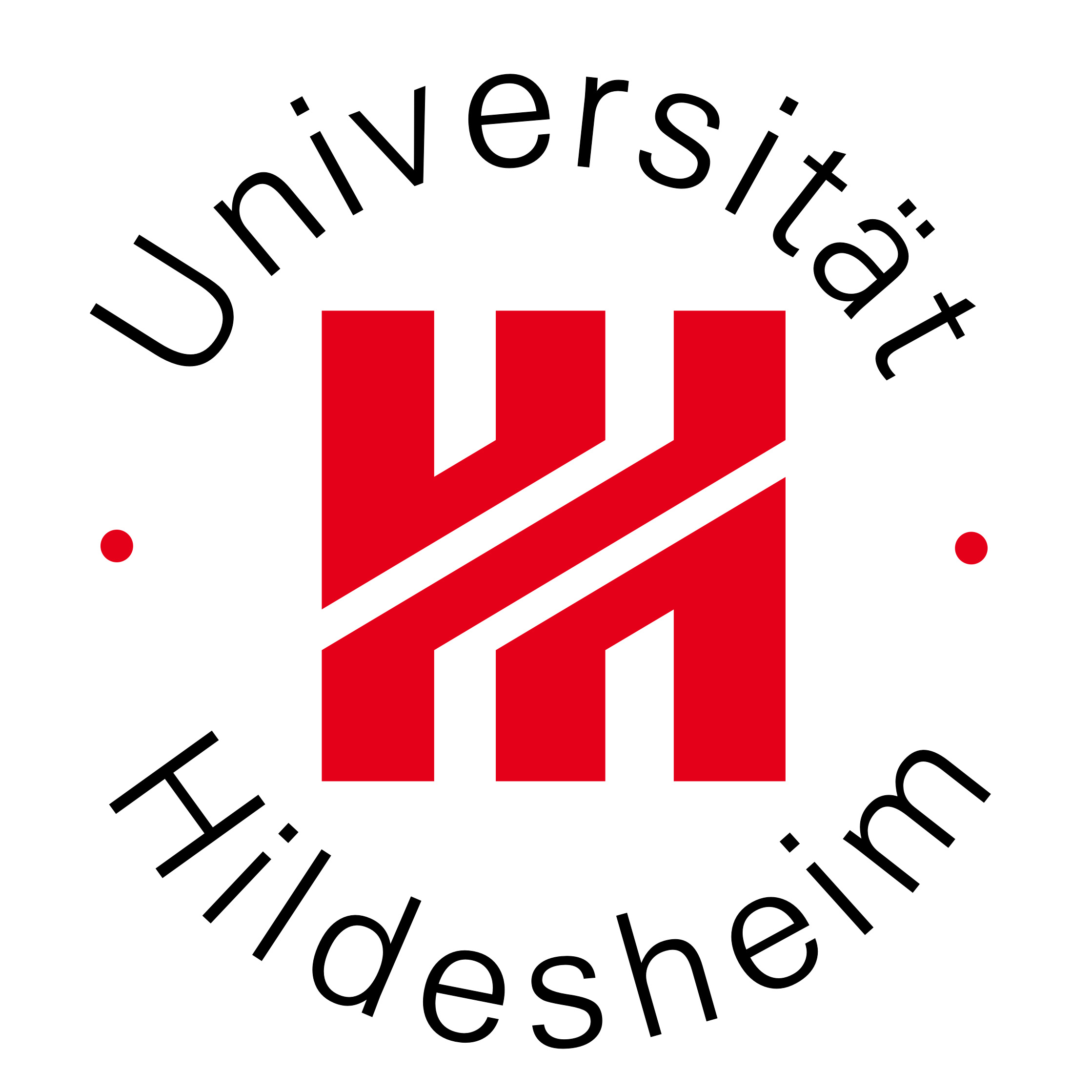}
\vspace{2cm}


{ \Large \bfseries Reinforcement Learning Approach to Active}\\  
\vspace{0.5cm}
{ \Large \bfseries Learning for Image Classification}\\ 
\vspace{0.5cm}
\vspace{2cm}


\begin{minipage}{0.49\textwidth}
\begin{flushleft} \large
\emph{Author:}\\
Thorben \textsc{Werner} \\ 
279870 \\ 
\end{flushleft}
\end{minipage}
~
\begin{minipage}{0.46\textwidth}
\begin{flushright} \large
\emph{Supervisors:} \\
Prof. Dr. Dr. Lars \textsc{Schmidt-Thieme} \\ 
Mohsan \textsc{Jameel} 
\end{flushright}
\end{minipage}\\
\vspace{2cm}


{ \today}\\ 
\vspace{1cm}

{ \large \bfseries Thesis submited for}\\  
\vspace{0.3cm}
\textsc{\Large Master of Science in IMIT - Angewandte Informatik}\\
\vspace{1cm}
\textsc{\large Wirtschaftsinformatik und Maschinelles Lernen}\\ 
\vspace{0.3cm}
\textsc{\large Stiftung Universität Hildesheim}\\ 
\vspace{0.3cm}
\textsc{\large Universitätsplatz 1, 31141 Hildesheim}\\ 
\vspace{0.3cm}

\vfill 

\end{titlepage}

\setcounter{secnumdepth}{1}


\noindent \textbf{Statement as to the sole authorship of the thesis:}
\vspace{0.4cm}
\\Reinforcement Learning Approach to Active Learning for Image Classification.\\[1mm]
I hereby certify that the master's thesis named above was solely written by me and that no assistance was used other than that cited. The passages in this thesis that were taken verbatim or with the same sense as that of other works have been identified in each individual case by the citation of the source or the origin, including the secondary sources used. This also applies for drawings, sketches, illustrations as well as internet sources and other collections of electronic texts or data, etc. The submitted thesis has not been previously used for the fulfillment of degree requirements and has not been published in English or any other language. I am aware of the fact that false declarations will be treated as fraud.
\vspace{7cm}

\today, Hildesheim

\thispagestyle{empty}
\setcounter{tocdepth}{2}
\newpage


\begin{abstract}
Machine Learning requires large amounts of labelled data to fit a model. Many datasets are already publicly available, nevertheless forcing application possibilities of machine learning to the domains of those public datasets.
The ever-growing penetration of machine learning algorithms in new application areas requires solutions for the need of data in those new domains. This thesis works on active learning as one possible solution to reduce the amount of data that needs to be processed by hand, by processing only those datapoints that specifically benefit the training of a strong model for the task. \\
A newly proposed framework from \cite{howToActiveLearn} for framing the active learning workflow as a reinforcement learning problem is adapted for image classification and a series of three experiments is conducted. Each experiment is evaluated and potential issues with the approach are outlined. Each following experiment then proposes improvements to the framework and evaluates their impact. After the last experiment, a final conclusion is drawn, unfortunately rejecting this work's hypothesis and outlining that the proposed framework at the moment is not capable of improving active learning for image classification with a trained reinforcement learning agent. 
\vfill
\end{abstract}

\newpage

\tableofcontents
\listoffigures
\listoftables
\lstlistoflistings
\newpage
\pagenumbering{arabic}

\chapter{Basics}

\section{Machine Learning}
Machine Learning (ML) is one of the two major sub-fields of artificial intelligence, the other one being formal reasoning. ML is the science of learning abstract rules from data without human supervision. For this, one needs to decide on one target variable $y = \{ 1, \cdots, C \}$ that will be subject of the prediction of the model. Next, a dataset of features \textbf{x} $= \{ x_1, \cdots, x_M \}$ is collected that will serve as an input to the model. The result is a collection of observations and corresponding values of the target variables $D = \{($\textbf{x}$^{(1)}, y^{(1)}), \cdots, ($\textbf{x}$^{(N)}, y^{(N)}) \}$. \\
The learning process is characterized by the model, which is defined by its parameters $\theta$, the employed loss function $L(D, \theta)$ and the learning procedure. Generally, the learning procedure tries to optimize $\theta$ so that the loss function is minimal. At this minimum the model is said to have optimal predictive performance. \\
The most recent form of rule abstraction is the organisation of layers that learn increasingly complex dependencies, starting with very simple abstractions in the first layer. This architecture is called a deep neural network (NN). \\[2mm]
This section also provides a list of machine learning terms that will not be described in detail later in the thesis.

\paragraph{Gradient Descent}
Optimization technique to train many different machine learning models. Uses the derivative of the loss function with respect to the parameters of the model to update the parameters in small steps. Ideally minimizing the error the model is making in the process and fitting it to the task at hand.

\paragraph{(Leaky) ReLU}
Rectified Linear Unit. Non-linear function used in Neural Networks. Does not change any non-negative inputs, but truncates all values below 0 to be exactly 0. The leaky version does not truncate values below 0, but rather discounts them with a factor $a$
\begin{figure}[H]
    \centering
    \includegraphics[width=\textwidth]{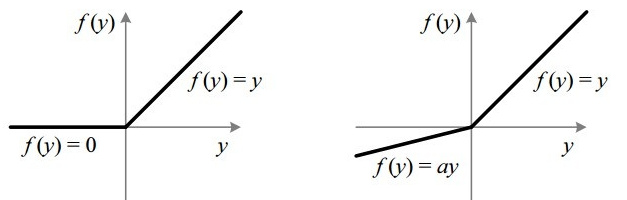}
    \caption{Visualization of ReLU (left) and Leaky ReLU (right) \hspace{2mm}Source: \cite{reluImg}}
\end{figure}

\paragraph{Softmax}
Exponential function that is used for many different purposes in machine learning. Normalizes a given vector of values to be non-negative and to sum to 1, turning it into a probability distribution. Softmax also emphasizes high values and suppresses low values in the process.

\section{Neural Networks}
\begin{figure}[h]
    \centering
    \includegraphics[width=0.7\textwidth]{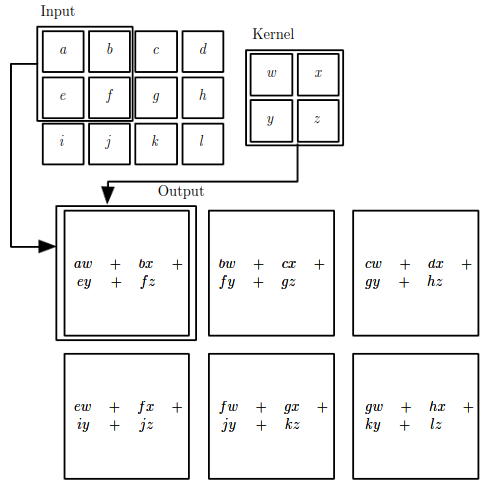} 
    \caption{Visualization of a convolution (\cite{bookDeepLearning} p. 330)}
    \label{fig:convExample}
\end{figure}

Deep Neural Networks consist of layers of ``artificial neurons''. Each neuron creates a linear combination of its inputs $a = Wx$, with $W$ being the neurons weight matrix and $x$ being the vector of inputs. Before the neuron passes this linear combination to the next layer, a non-linear activation function $\phi(a)$ is applied to introduce non-linearity into the network. The collection of all weight matrices defines the model parameters $\theta$.\\
Each linear combination followed by an activation function represents one rule that is learned by the network. The first layer now learns simple rules on the raw data. The second layer, being fed the outputs from the first layer, learns a linear combination of the rules from the first layer and so on. This architecture is able to learn higher and higher abstractions of the data with increasing depth of the network.\\[1mm]
Neural networks have many different archetypes that are specialized to solve certain problems. The networks that were described up to this point were fully connected (dense) networks. An other archetype is the convolutional neural network (CNN) that was developed by Yann LeCun et. al. \cite{origConvNet} to efficiently process images. Each layer of this type defines one or multiple sliding windows, called filters, that are moved across the image combining a certain number of input pixels into a single pixel of higher abstraction. This combination again is a linear combination usually followed by an activation function and is defined by the weight matrix (kernel) of the filter. A visualization of this process can be found in figure \ref{fig:convExample}.

\section{Active Learning}
Burr Settles, the author of a heavily cited active learning survey \cite{alSurvey} and follow-up book \cite{alBook}, describes active learning (AL) as follows: ``Active learning [...] is a subfield of machine learning and, more generally, artificial intelligence. The key hypothesis is that if the learning algorithm is allowed to choose the data from which it learns — to be “curious,” if you will — it will perform better with less training.''(\cite{alSurvey} p.4)\\
A strong model that is trained on very few datapoints is especially valuable if the cost of acquiring labels is high, or if the model generally needs an immense quantity of random datapoints to learn effectively.\\
The basic active learning framework consists of four components: the unlabeled dataset $U$, the labeled dataset $L$, the model, parameterized by $\theta$, and a sampling strategy $\phi(p(U,\theta))$. 
\begin{lstlisting}[escapeinside={(&}{&)}, caption={Generic Active Learning Workflow}, captionpos=t, label={alg:activeLearning}]
INPUT: Unlabeled Dataset (&$U$&), Model (&$\theta$&), Sampling Strategy (&$\phi$&), Budget
(&$L$&) = {} # Start with an empty labeled set
i = 0
while i < Budget:
    # Retrain model 
    fit((&$\theta$&), (&$L$&))
    # Find most promising datapoint to label
    (&$x^*$&) = (&$\phi$&)(p((&$\theta$&), (&$U$&)))
    # Obtain label from oracle and add to dataset
    (&$L$&) = (&$L \cup (x^*, y^*$) &)
    (&$U$&) = (&$U \setminus x^*$ &)
    i += 1
\end{lstlisting} 
One iteration of AL is performed by calculating the predictions of the model on the unlabeled dataset $P_\theta(\hat y \mid U)$. Then, based on that information the sampling strategy $\phi$ picks a point $x^\star \in U$ that maximizes the expected improvement of the model when added to the labeled dataset. After $x^\star$ is selected, an oracle provides a label for the datapoint, it is added to the labeled set $L$ and the model is retrained.

\section{Reinforcement Learning}\label{sec:basicRL}
Reinforcement Learning is the field of optimizing an agent, which acts in an environment, based on a reward signal. The agent tries to maximize the reward by assessing the state that it is currently in and estimating the expected reward of possible actions from this point. Most of the time this is done by using the framework of a Markov Decision Process (MDP) to model the environment in form of it's state space $S$, possible actions $A$ and the reward function $r$, as described by Sutton and Barton (\cite{bookRL} pp. 47-68).
The interactions of agent and environment take place during discrete timesteps $T \in \mathbb{N}$ and are subject to a state transition function $p$. The resulting tuple ($S, A, T, p, r$) fully descibes the dynamic of the MDP.\\
Sutton and Barton characterize reinforcement learning (RL) as the third paradigm of machine learning, besides supervised and unsupervised learning (\cite{bookRL} p. 2). This is due to the additional challenges of RL that do not appear in traditional methods of ML. The most prevalent one being the need to balance exploration and exploitation in the training process. The agent should always pick the action that yields the maximum expected reward (exploitation), but in order to discover the true value of each action in different situations the agent needs to explore unknown or sub-optimal actions (exploration).\\ [1mm]
This work focuses on Q-Learning. Formally, an agent, represented by a neural network $\pi$, also called Deep Q Network (DQN), receives a state $s_i$ and predicts the so-called Q-Value for every possible action $a_i$. This Q-Value $\hat Q_\pi (s_i, a_i)$ indicates the quality of $a_i$ for that given state $s_i$. Apart from exploration scenarios, the agent always greedily picks the action with the highest Q-Value.\\[1mm]
The neural network is optimized via gradient descent with Mean Squared Error (MSE) as loss function, resulting in a standard regression problem displayed in Equation \ref{eq:bellmanTarget}. The target $y_i$ is given by the Bellman-Equation (\cite{bookRL} p. 157) and $i = 1, \cdots, N$ indicates the iterations during training.
\begin{equation}\label{eq:bellmanTarget}
    \pi^* = \underset{\pi}{argmin} \hspace{1mm} \frac{1}{2N} \sum\limits_{i=1}^N \| \hat Q_\pi (s_i, a_i) - y_i \|^2_2 
\end{equation}
\begin{equation*}
    y_i = r_i + \gamma \hspace{1mm} \underset{a}{max} \hspace{1mm} \hat Q_\pi(s_{i+1}, a)
\end{equation*}
An update is performed after an interaction of the agent with the environment and both a reward $r_i$ and the follow-up state $s_{i+1}$ have been obtained. $\hat Q_\pi(s_i, a_i)$ is updated to reflect the received reward plus the best possible action in the follow-up state. The discounting factor $\gamma$ encourages the agent to find rewards early, as later ones are discounted by $\gamma^k$ after $k$ steps.
The basic algorithm is summarized in Algorithm \ref{alg:simpleRL}
\newpage
\begin{lstlisting}[escapeinside={(&}{&)}, caption={Reinforcement Learning Algorithm}, captionpos=t, label={alg:simpleRL}]
INPUT: Environment (&$Env$&), Agent (&$A_\pi$&), NumInteractions
i = 0
while i < NumInteractions:
    done = False # indicates when a game terminated and the environment needs to be reset   
    (&$s_i$&) = Env.reset() # Reset the environment and receive initial state
    while not done:
        # Predict Q values and determine action 
        Q, (&$a_i$&) = (&$A_\pi$&).predict((&$s_i$&))
        # Use action to interact with the environment
        (&$s_{i+1}$&), (&$r_i$&), done  = (&$Env$&).step((&$a_i$&))
        # Use new state (&$s_{i+1}$&) to update the agent with equation (&\ref{eq:bellmanTarget}&)
        (&$A_\pi$&).update((&$s_i$&), (&$a_i$&), (&$s_{i+1}$&), (&$r_i$&))
        (&$s_i$&) = (&$s_{i+1}$&)
        i += 1
\end{lstlisting}

\chapter{Introduction}
Machine Learning (ML) starts to penetrate all areas of public and private life. This development gives rise to an ever increasing need for training data to support all the new ML models.
Since many models also require a ground truth to be attached to every datapoint, creating a new dataset becomes increasingly difficult and costly. One not only needs to collect high amounts of data, but also needs to label each datapoint by hand. Companies like Google started to provide labeling services for machine learning datasets, that consist of a big workforces of human labeling experts \cite{googleLabeling}. For individuals, however, no good solution exists. 
To help anyone that needs to create a new labeled dataset, one can either try to automate the process, or reduce the amount data that needs to labeled by hand.
One way to reduce the amount of data that needs to be labeled is to employ active learning. The goal is to train a strong model for a new ML task on as few datapoints as possible, requiring the datapoints to be carefully selected in order to maximize their benefit for model training.
Even though many different sampling strategies for AL have existed for a long time, the current surge in reinforcement learning introduced a new angle of improvement for active learning.
On a fundamental level, AL can be framed as a RL problem, namely a sequence of decisions (which datapoints to pick) leads to a final result that can be evaluated in quality (performance of the model), therefore emitting a reward signal that can be used by a RL agent.\\
This thesis picks up the work of Fang et. al. \cite{howToActiveLearn} that proposes a  formulation of stream-based AL as a RL problem and evaluates the performance of the trained agent on a Named Entity Recognition (NER) task. The paper will be described in detail in Section \ref{sec:howToActiveLearn}. \\[1mm]
In a series of three experiments, this thesis will apply the work of \cite{howToActiveLearn} to image classification, propose several improvements to the training framework and the formulation of the RL problem and evaluate on the performance of the RL agent compared to traditional AL sampling strategies. \\
After a methodology section in Chapter \ref{chap:method}, the experiments are described in Chapter \ref{chap:exp1}-\ref{chap:exp3}. A final conclusion and an extensive list of possible future works are given Chapter \ref{chap:conclusion} and \ref{sec:futureWork}.

\chapter{Related Work}
\section{Active Learning}
The field of active learning has been traversed by multiple surveys already. A central piece is the ``Active Learning Literature Survey'' by Burr Settles \cite{alSurvey}, which includes a definition of the AL problem, an exhaustive list of sampling strategies, different AL frameworks and many practical considerations. Also, more recent works like the ``Survey on instance selection for active learning'' by Fu et. al. \cite{newAlSurvey} are available, but don't add additional value.

\subsection{Active Learning for Image Classification}
The work of Joshi et. al. \cite{rwAL1} is a typical example of active learning for image classification. The authors use uncertainty sampling to select images from an unlabeled pool to train SVMs. The combination of uncertainty sampling and SVM models is the most common in this sub-field of AL. A more widespread study was conducted by Tuia et. al. \cite{rwAL2}, who created a survey similar to \cite{alSurvey} but limited to applications for satellite image data. Hoi et. al. \cite{rwAL3} explore the use of batch AL in the fields of medical image analysis, again using SVMs as predictive model for medical datasets like the UCI breast cancer dataset.

\subsection{Active Learning with Neural Networks}
As mentioned above, using neural networks in an active learning framework is quite uncommon. The work of Wang et. al. \cite{rwAL4}, however, uses a CNN for face recognition and object detection. They state that one of the problems is that ``Most AL methods pay close attention to model/classifier training. Their strategies to select the most informative samples are heavily dependent on the assumption that the feature representation is fixed.''(\cite{rwAL4} p.1) The authors hint at the fact that most models that are used in AL have a fixed feature representation (mainly SVMs), but CNNs optimize their feature selection jointly with the decision boundary. To overcome this issue Wang et. al. propose to not only add samples of low confidence (high informativeness) to the labeled set, but also complement it with high confidence samples to quickly grow the training set of the CNN.\\
Gal et. al. \cite{rwAL5} use the Bayesian equivalent of CNNs to reintroduce a proper model uncertainty that is missing with conventional CNNs, but is important for active learning. To prove this importance the authors compare the performance of active learning with conventional CNNs against Bayesian CNNs and conclude that a proper uncertainty measure for the classification model at hand is highly beneficial for standard AL methods.

\section{Reinforcement Learning}
The combination of reinforcement learning and active learning is extremely rare in the current state-of-the-art. One of two works is ``Learning to active Learn: A deep reinforcement learning approach'' by Fang et. al. \cite{howToActiveLearn}. This paper serves a baseline paper and will be discussed in detail below. \\
The second paper is ``Active learning for reward estimation in opposite reinforcement learning'' by Lopes et. al. \cite{rwRL1} which is combining reinforcement learning and active learning in the inverse way this work intents to. The authors use active learning to aid the learning process of a RL agent instead of enhancing active learning with a RL agent. \\
Considering the absence of compatible works in the field, this section will instead introduce three well known papers that provide important pieces for the employed training framework.

\subsection{Human-level control through deep reinforcement learning (2015)}
The most cited paper in the field of reinforcement learning on Google Scholar probably is the starting point for most reinforcement learning endeavors. The authors Mnih et al. \cite{mnih2015} compiled an excellent guide for successful reinforcement learning frameworks and training procedures. \\
Establishing the essential components for RL, like a memory buffer, a separate target network for Q-Learning and the use of a deep convolutional architecture. \\
In addition, Mnih et. al. provide a comprehensive list of hyperparameters that pose a solid configuration for a wide range of RL frameworks.

\subsection{Prioritized Experience Replay (2016)}
The uniform sampling of memories in RL is commonly known to be highly inefficient. 
To address this issue Schaul et. al. propose a technique called ``Prioritized Experience Replay'' \cite{prioReplay} that assigns a higher sampling probability to more informative samples (indicated by the respective loss of that sample). 
This vastly improves scenarios in which a small amount of important samples with non-zero reward is buried under a mass of ``failures'' with zero rewards, causing a need for excessive amounts of training iterations for an agent to learn a useful policy. (\cite{prioReplay} pp. 2-3)
The authors apply their sampling technique to the experiments of \cite{mnih2015} and show significant improvements on various games, compared to traditional uniform memory sampling.

\subsection{Deep Reinforcement Learning with Double Q-Learning (2016)}
The introduction of a second Q-network into the Q-learning framework was already proposed and implemented  by previous papers, however, van Hasselt et. al. \cite{ddqn} properly decoupled the target and action network in during optimization to further stabilize the training of reinforcement learning agents. 
Motivated by a study on overestimation errors, the authors propose to use the online network to evaluate the greedy policy while the target network estimates its value. 
The paper includes empirical results that show that the Double-DQN (DDQN) successfully overcomes the overestimation bias that normal DQN architectures inherit.

\chapter{Methodology}\label{chap:method}

\section{Active Learning}\label{sec:al}
The following sections describe the different sampling strategies $\phi$ that can be used in Algorithm \ref{alg:activeLearning}, and how active learning is applied to image classification. 

\subsection{Uncertainty Sampling} \label{sec:uncertaintySampling}
\begin{figure}[h]
    \begin{center}
        \includegraphics[width=\textwidth]{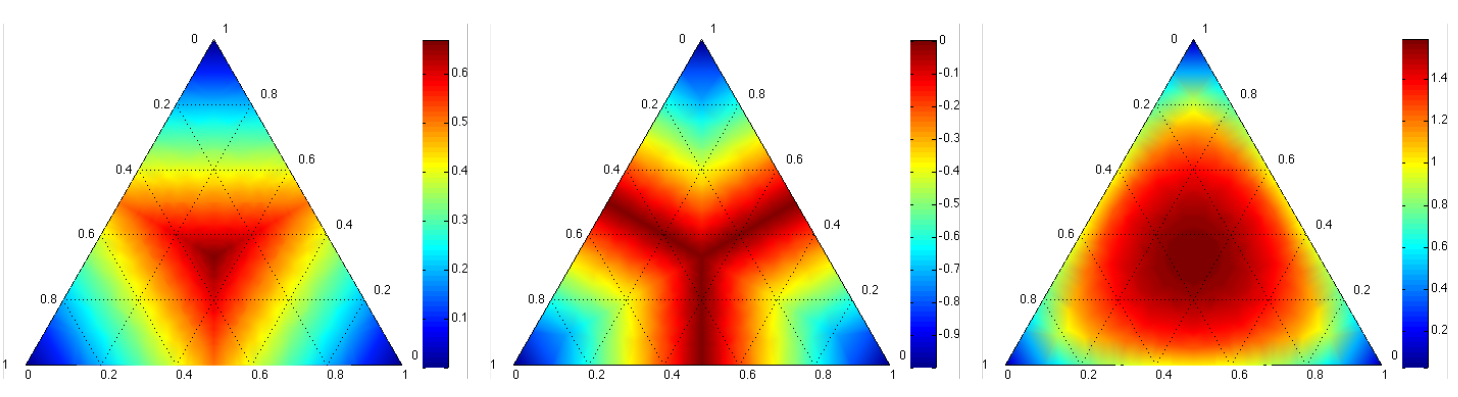}
    \end{center}
    \caption*{\hspace{-5mm} (i) least confident \hspace{17mm} (ii) BvsSB \hspace{25mm} (iii) entropy}
    \caption{Most informative areas in a 3 class problem according to the sampling strategies (\cite{alSurvey} Fig. 5)}
    \label{fig:samplingStratz}
\end{figure}

The most common sampling strategy is to pick the instance from the unlabeled set $U$ for which the model is most ``unsure''. This is straightforward for probabilistic models as they provide their certainty in the output. Different metrics can be employed to evaluate the uncertainty (\cite{alSurvey}  pp. 12-15):\\
(i) the least confident value among all of the highest class probabilities
\begin{align*}
    x^\star = \underset{x \in U}{argmax} \hspace{2mm} 1 - P_\theta(\hat y\mid x) \\
    with \hspace{4mm} \hat y = \underset{y}{argmax} \hspace{2mm} P_\theta(y\mid x)
\end{align*}
(ii) the difference between the highest $\hat y_1$ and second highest $\hat y_2$ class probability, also known as ``Best vs Second Best'' (BvsSB)
\begin{align*}
    x^\star = \underset{x \in U}{argmin} \hspace{2mm} P_\theta(\hat y_1 \mid x) - P_\theta(\hat y_2 \mid x)
\end{align*}
(iii) the Shannon entropy over the class probabilities
\begin{equation*}
    x^\star = \underset{x \in U}{argmax} \hspace{1.5mm} - \sum\limits_{c=1}^C P_\theta(y_c\mid x) log P_\theta(y_c\mid x)
\end{equation*}
\newpage
A visual comparison of the sampling strategies can be found in Figure \ref{fig:samplingStratz}. The graphs picture a three class scenario where the position within the triangle represents the predicted probability for each of the three classes for a given instance. The color represents the grade of information that instance provides to the model if added to the labeled set.
Each strategy shares the property that the most informative samples can be found in the middle of the triangle, where the probability for each class is leveled at $\frac{1}{3}$, and the least informative samples are at the corners, where one class has a probability of 1 and the rest has 0. Considering the edges, however, they differ significantly from one another. Strategy (i) and (iii) both are very centralised, favoring those instances that produce a low overall confidence in the prediction. Strategy (ii) only focuses on the two highest probabilities, favoring a high uncertainty between those two. By doing this, the BvsSB strategy solves a common problem of (i) and (iii). In situations with a large number of labels, the probability distribution $P_\Theta(\hat y \mid x)$ tends to flatten out towards all the unlikely labels. This lowers the confidence of more likely labels compared to the mass of unlikely labels as well as causing a high entropy for every sample $x \in U$. Thus hindering the performance of the least confident (i) and entropy (iii) strategy.

\newpage
\section{Learning how to Active Learn}\label{sec:howToActiveLearn}
Starting point for this work is the paper ``Learning how to Active Learn: A Deep Reinforcement Learning Approach'' a paper from Meng Fang, Yuan Li and Trevor Cohn \cite{howToActiveLearn}. Fang et. al. trained a Named Entity Recognition (NER) model on a limited set of labeled data that was curated by an AL procedure. They motivated their work with the profound usecase of AL inside a low-resource language. Contrary to existing methods, Fang et. al. did not use well known heuristics for the AL sampling (see Sec. \ref{sec:al}) but trained a reinforcement learning agent to do the sampling instead. The authors use a high resource language (English) to learn a sampling policy, which then can be transferred to a low-resource language setting (German, dutch, Spanish).\\[1mm]
To train the RL agent Fang et. al. reformulated the active learning problem into a Markov decision process (MDP) by presenting the agent a single unlabeled datapoint and defining the action space as \{\textit{Label}, \textit{NotLabel}\}.
This policy can be optimized with reinforcement learning to pick those datapoints that maximize the expected improvement in model performance.
\begin{lstlisting}[escapeinside={(&}{&)}, caption={Learn an active learning policy (\cite{howToActiveLearn} Alg. 1)}, captionpos=t, label={alg:baselineALRL}]
INPUT: Unlabeled Dataset (&$U$&), Budget (&$B$&), NER Model (&$\theta$&), RL Agent (&$Q_\pi$&), Memory Buffer (&$M$&)
for episode in [1, ... , N]:
    (&$L$&) = {} # Start with an empty labeled set
    (&$\theta$&) = random
    for i in [1, ..., |D|]:
        construct state (&$s_i$&) using (&$x_i$&)
        (&$a_i$&) = (&$\underset{a}{argmax} \hspace{1mm} Q_\pi (s_i, a)$&) # obtain agent's decision
        if (&$a_i$&) = 1:
            obtain annotation (&$y_i$&)
            (&$L$&) = (&$L$&) (&$\cup$&) ((&$x_i$&), (&$y_i$&))
            fit (&$\theta$&) to (&$L$&)
        generate reward (&$r_i$&) using validation data
        if |D| = (&$B$&):
            store transaction (&$M$&) = (&$M$&) (&$\cup$&) (&$(s_i, a_i, r_i, DONE)$&)
            break
        construct new state (&$s_{i+1}$&)
        store transaction (&$M$&) = (&$M$&) (&$\cup$&) (&$(s_i, a_i, r_i, s_{i+1})$&)
        sample minibatch (&$(s_j, a_j, r_j, s_{j+1}) \sim M$&)
        perform gradient descent according to Eq. (&\ref{eq:bellmanTarget}&)
\end{lstlisting}

\paragraph{Note} Algorithm \ref{alg:baselineALRL} is meant to be as close as possible to it's source and therefore does deviate stylistically from other algorithms presented in this thesis. \\[1mm]
The paper details an implementation for the state representation and reward function. It proposes a state at the i-\textit{th} time step $s_i$ that consists of the concatenation of an embedding of the considered datapoint $x_i$, an embedding of the model marginals $p_\theta(y \mid x_i)$ and a confidence measure $C$ based on the model. The reward is calculated based on the improvement of the NER model on a validation set. For this, F1-Score is used as metric.
The construction of the deep Q-network (DQN) and the training procedure follows Mnih et. al. \cite{mnih2015}. The authors use convolutional layers to embed the considered datapoint $x_i$ and the marginals $p_\theta(y \mid x_i)$ and a single layer DQN to predict the Q-values. During training, experience replay \cite{mnih2015} and stochastic gradient descent with batch size 32 are used. The NER model is a standard linear chain CRF (\cite{CRF})\\
This thesis omits the presented graphs, as they display specific cross-lingual experiments, which are not comparable to any of the following work. Generally, Fang et. al. showed a clear improvement of the trained AL-agent over the classic entropy based sampling strategy in all experiments.

\subsection{Active Learning for Image Classification}
Conceptually, the AL workflow does not change when applied to image classification. However, the used model $M$ tends to be more complex. Since up to one model per picked datapoint is trained, AL usually uses easy to train models like SVMs. To enable the use of an arbitrary image dataset, a CNN is employed in the further experiments of this work.\\
This makes it hard to supplement the heuristic $\phi$ with information besides the final prediction $p_\theta(\hat y \mid x)$. Furthermore, the training procedure for CNNs inherits a high amount of variance. Different initializations require varying amounts of iterations to reach a certain level of performance, which in turn might even be unobtainable for other initializations.

\newpage
\section{Reinforcement Learning}
A reinforcement learning approach is characterized by three components: (i) a neural network that acts as an agent, (ii) the environment in which that agent operates and (iii) the learning algorithm used to update the agent.

\subsection{Active Learning Environment}\label{sec:ALEnv}
The proposed environment for the active learning game conforms to the OpenAI Gym \cite{openAiGym} interface. It exposes the functions 
\begin{lstlisting}
def reset(): return state
def step(action): return newState, float reward, bool done, info
\end{lstlisting}
and the fields
\begin{lstlisting}
int actionSpace
int[] stateSpace
\end{lstlisting}
Apart from the usual functions, the environment needs to maintain the labeled and the unlabeled datasets as well as the image classification (IC) network for the active learning workflow.\\
The internal state of the environment is defined by a list of datapoint IDs that play the role of the current position of the agent in conventional RL environments. Based on these IDs, whose respective datapoints are fed into the image classification model, the current state vector can be calculated.
The exact nature of the state vector depends on the experimental setup and is detailed in the respective chapters below.\\
The action space is defined by a number of categorical actions. All actions but the last add an image to the labeled dataset and trigger the training process of the image classification model. The last action always is a special ``pick no image'' option. When chosen, a random image from the current state is replaced with a new one and no training of the IC model takes place.\\
The default reward of each action is 0, deviating only if reward shaping is enabled or the game has ended. In those two cases the reward is equal to the improvement in F1 score of the IC model on a held-out validation set compared to the last time a reward was generated.\\
To keep the running time feasible, model training is heavily sanctioned. The validation error is measured on a reduced validation set of 1000 samples and an aggressive early stopping is employed which stops as soon as the validation error is not monotonically decreasing. This has potential negative effects on the reward signal. Even though the training set (the environments labeled dataset) barely changes in most cases and therefore the F1-Score should deviate only slightly, the reward was fluctuating heavily. To balance this, the F1-Score $F_i$ was internally modeled by a moving average 
\begin{equation*}
    F_i = \alpha F_{i-1} + (1-\alpha)F_i \hspace{5mm} with \hspace{5mm} \alpha = 0.7 
\end{equation*}
Each game in this environment is initialized with an initial labeled set of 5 images per class. In active learning this is called the seed set.
\begin{figure}[H]
    \centering
    \includegraphics[width=\textwidth]{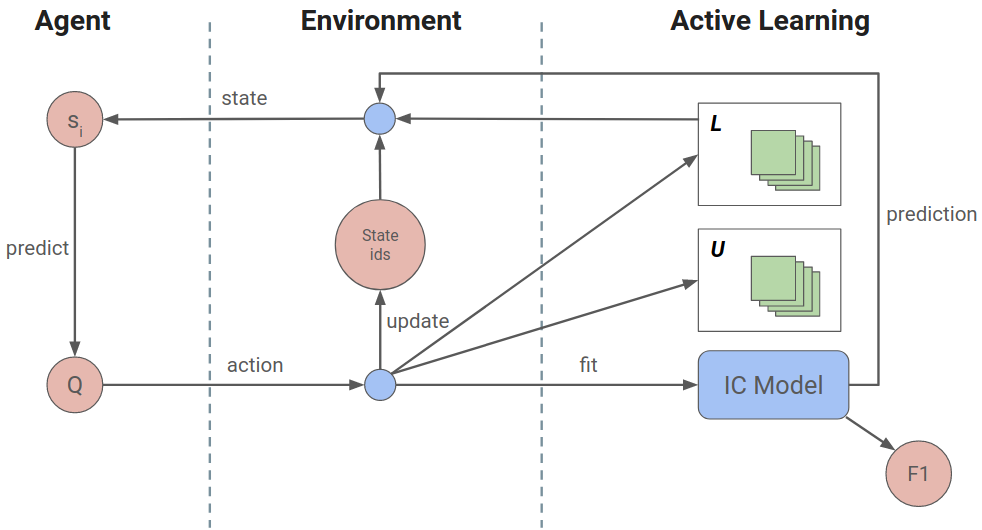}
    \caption{Architecture of the RL environment}
    \label{fig:envArch}
\end{figure}

\subsubsection{Additional Parameters of the active learning environment}\label{lst:alEnvFields}
\begin{itemize}
 \item model - The image classification model that is used
 \item dataset - A tuple of training and validation data
 \item {\color{blue} int} initialPointsPerClass - Number of labeled images per class at game start
 \item {\color{blue} bool} rewardShaping  - Determines, if a reward should be generated after every interaction, or only at the end of a game 
 \item {\color{blue} int} budget - The number of images that can be added to the labeled set before the environment resets
 \item {\color{blue} int} maxInteractionPerGame - The number of interactions after which a game terminates even if the budget is not exhausted
\end{itemize}

\newpage
\subsection{DQN / DDQN}
The agent in reinforcement learning is modeled by a neural network. The agent tries to act intelligently by assessing his current situation and the potential value of actions from this point. The resulting behavior is also known as a policy. A well known approach to do this is to estimate the so called Q-Value of an action in a given state $s_i$ with a Deep Q Network (DQN). The Q-Value models the expected sum of future rewards when following the current policy. Practically, this resolves into a regression problem, where the agent network takes a state as input and outputs one estimated Q-Value for each possible action.\\
To determine an action $a_i$ from the estimates one can simply take the action with the highest Q-Value. This behavior is called a greedy policy and is usually applied during evaluation of the agent. During training however, the agent needs to resort to some form of exploration in order to arrive at optimal solutions. A behavior that is impossible with a greedy policy. To introduce variance, either a $\epsilon$-greedy policy can be used, where the agent takes a random action with a probability of $\epsilon$, or the Q-values are passed through a softmax function to convert them into a probability distribution, from which an action can be sampled.\\
\begin{equation}\label{eq:softmaxGreedy}
    a_i \sim Cat(A, P_\theta)
\end{equation}
\begin{equation*}
    P_{\theta, k} = \frac{exp(Q_k(s_i) / \tau)}{\sum\limits_{j=1}^A exp(Q_j(s_i) / \tau)} \hspace{3mm}, k = 1, \cdots, A
\end{equation*}
The parameter $\tau$ is called the greed parameter and controls the decisiveness of the distribution, singling out the highest value when approaching 0.\\
To optimize the agent network, Alg. \ref{alg:simpleRL} can be used. Following well known papers, this procedure is extended with a Memory Replay Buffer \cite{mnih2015} and a Double DQN (DDQN) agent \cite{ddqn}.

\paragraph{Memory Replay} 
Instead of using only the most recent interaction with the environment (namely the tuple ($s_i, a_i, s_{i+1}, r_i$) from Alg. \ref{alg:simpleRL}), every interaction is stored in a memory buffer from which a minibatch of interactions can be retrieved for agent training. Using an interaction to fit the agent does not remove it from the memory, but the memory buffer usually has a maximum length. This gradually removes old and potentially less relevant interactions, while still raising the data efficiency by enabling interactions to be used multiple times. This thesis will reside on uniform sampling of interactions and not look into prioritized replay \cite{prioReplay}.

\paragraph{DDQN}
The double DQN architecture was proposed by van Hasselt et al. \cite{ddqn}. The authors tackle the common problem of overestimations in deep Q-networks by proposing an architecture with two Q-networks. One primary network to define the policy and one secondary network that is used in the formulation of the regression target.\\
General Equation from Section \ref{sec:basicRL} Eq. \ref{eq:bellmanTarget}
\begin{equation*}
    y_i^{DQN} = r_i + \gamma \hspace{1mm} \underset{a}{max} \hspace{1mm} \hat Q_\theta(s_{i+1}, a)
\end{equation*}
Decoupling of state selection and state evaluation (\cite{ddqn} Eq. 4)
\begin{equation*}
    y_i^{DQN} = r_i + \gamma \hspace{1mm} \hat Q_\theta \left( s_{i+1}, \hspace{1mm} \underset{a}{argmax} \hspace{1mm} \hat Q_\theta(s_{i+1}, a) \right)
\end{equation*}
Using the primary network to pick future actions and the secondary network $Q_{\theta'}$ to evaluate the state (\cite{ddqn} Eq. 4)
\begin{equation}
    y_i^{DDQN} = r_i + \gamma \hspace{1mm} \hat Q_{\theta'} \left( s_{i+1}, \hspace{1mm} \underset{a}{argmax} \hspace{1mm} \hat Q_\theta(s_{i+1}, a) \right)
\end{equation}
By separating the action selection and state evaluation during training, Hasselt et. al. were able to show consistent improvements over the results of Mnih et. al. \cite{mnih2015} The secondary network can be updated by simply switching the role of primary and secondary network repeatedly, or, following the original idea of a target network \cite{mnih2015}, by copying the weights from the primary network to the secondary network in a fixed interval $C$.

\section{Evaluation of an Active Learning Agent}\label{sec:alEval}
Since the classic sampling strategies for AL are simple to compute heuristics, the sampling strategy is usually presented the entire unlabeled dataset and picks the single most promising datapoint to be labeled. Then the classification model is retrained and the process is repeated until the budget is exhausted. (see Alg. \ref{alg:activeLearning}) \\
The evaluation scheme for this work will consist of one game in the AL environment with a budget of 800. After each added image the resulting F1-Score for the environment's image classification model is recorded and plotted into a graph of F1-Score vs Number of added Images.\\
Such a graph gives a good overview of the performance of an sampling agent in different stages of the AL process. Trivially, the higher the overall profile of the curve is the better a sampling strategy performs. The evaluation is performed multiple times and the results are averaged. As stated in Section \ref{sec:ALEnv}, the environment already tracks the F1-Score as a moving average. However, a second averaging with a sliding window of size 10 is applied to the curve before plotting to smooth the curves until they clearly separate and a trend becomes visible.

\section{Baselines}\label{sec:baselines}
\begin{figure}[H]
    \centering
    \includegraphics[width=.9\textwidth]{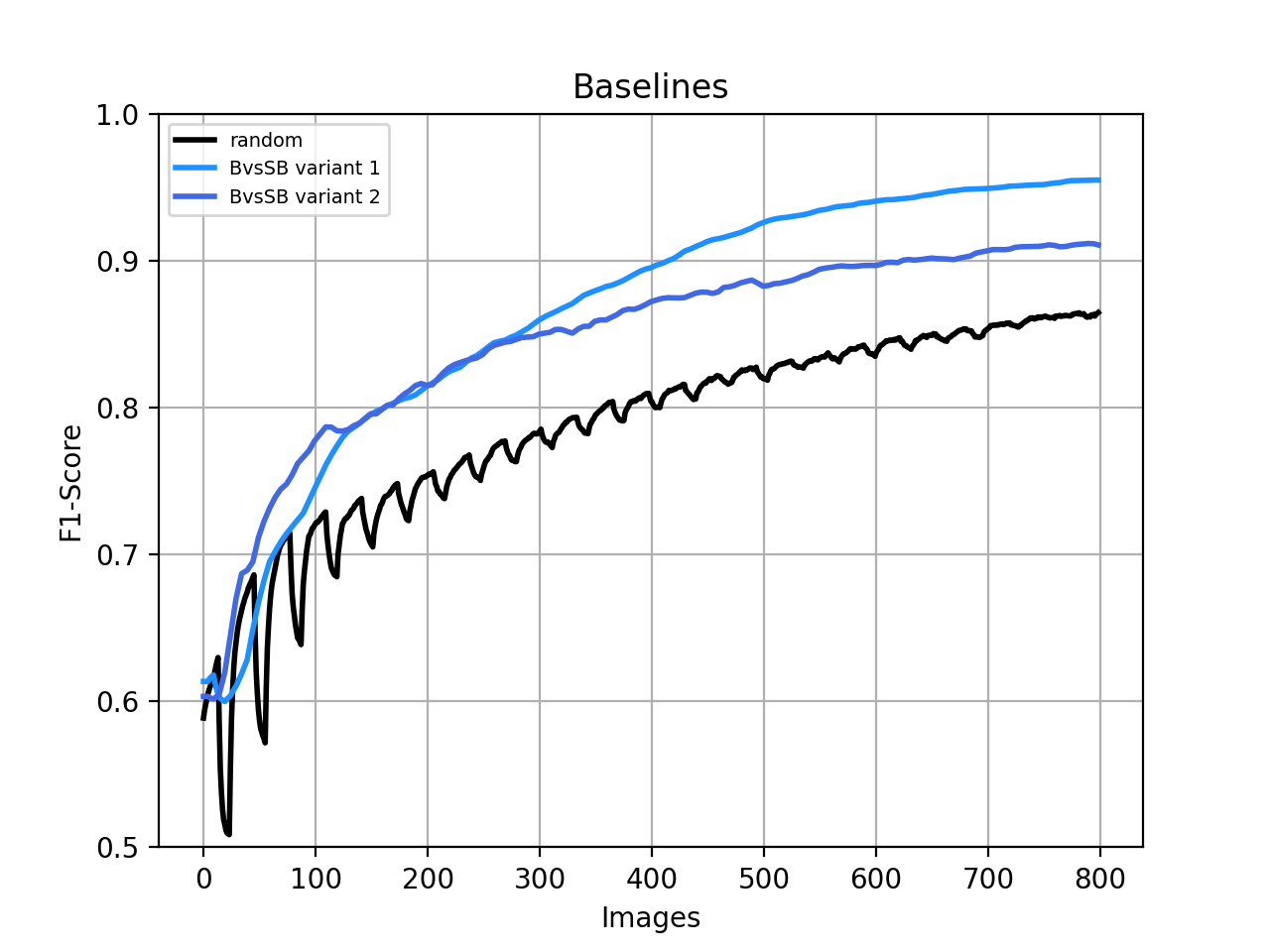}
    \caption{Baselines for AL agent evaluation}
    \label{fig:baselines}
\end{figure}

Each AL agent needs to compete against 3 baselines: first, against a random agent, who picks random images from the unlabeled dataset until the budget is exhausted and, second, against two variations of active learning with the BvsSB sampling strategy. A separate test of all mentioned sampling strategies of Sec. \ref{sec:uncertaintySampling} showed that the BvsSB is top-performing consistently.\\
Variant 1 of the BvsSB baseline complies to the default AL workflow and picks the single most promising image out of all unlabeled datapoints in each iteration.\\
Variant 2 is based on the workflow of Experiment 2, where only a small subset of the unlabeled dataset is presented at each iteration. During testing, this variant did not present any improvement over the random baseline and therefore was extended with an adaptive threshold that controls how many images are skipped before an image is chosen to be added. The baseline starts with a high initial threshold that slowly decays over time if no presented image generates a high enough BvsSB score. If an image exceeds the threshold, it is added to the labeled set and the threshold is reset to the initial high value.
The configuration of baseline variant 2 is as follows:
\begin{itemize}
    \item Number of presented images per interaction: 5
    \item Initial threshold: 0.8
    \item Threshold decay: 0.05
\end{itemize}
A comparison of the employed baselines can be seen in Fig. \ref{fig:baselines}.
The frequent drops in the random agent's curve are due to a phenomenon in the image classification model. By adding random images to the training set of the model, the IC model occasionally, due to an unfortunate distribution of training data, does not predict certain classes at all during evaluation and therefore negatively impacts the F1-Score. This phenomenon has reduced impact the bigger the training set gets.

\chapter{Experiment 1}\label{chap:exp1}
This and all following experiments use MNIST as dataset for the active learning process. The dataset is loaded through the \textit{keras.datasets.mnist} module and contains 60000 uint8-images of handwritten digits.
\paragraph{Note}
The following chapters will contain statements like ``the agent is presented a single image''. The phrases ``presented image'' and ``presented state'' will be used interchangeably and always are a simplification of ``the presented state, which is based on single or multiple images''.

\section{Experiment: Adaptation of the base paper to Image Classification}
To use the approach from the base paper for image classification some adjustments need to be made. Most notably, the classification model is switched from CRFs (\cite{howToActiveLearn} Chap. 4) to a convolutional neural network.\\
Additionally, some improvements are applied to the RL algorithm itself. Instead of using a single Q-network, Double Q-Learning \cite{ddqn} is employed. Instead of training a single Q-estimator, two networks are maintained simultaneously. This reduces the instability stemming from optimizing a moving target and fixes the overestimation problem described in \cite{ddqn}, where single network setups are overly optimistic due to estimation errors. \\
Following the base paper, the training procedure is a sequence of states generated from single images, presented to the agent. For each state the agent decides wether the corresponding image is to be labeled or not. After the action is taken, the interaction is stored in a memory buffer and the agent trained with a batch of memories. The state $s_i$ is consists of (i) entropy and BvsSB score of the  IC model's prediction $p_\theta(y \mid x_i)$ for the image, (ii) metrics on the IC model itself, namely the mean, standard deviation and norm of each layer and (iii) the average F1-Score of the current IC model.\\
The reward function is adopted from the base paper and consists of the improvement in F1-score of the image classification model on a validation dataset as discussed in Sec \ref{sec:ALEnv}. \\
In Experiment 1 and 2 the reward function can have two variants. One variant generates a reward after each action of the agent (reward shaping), while the other only generates a reward after a finished game (budget or maximum interactions exhausted).

\section{Execution and Results}
A full specification of the setup can be found in the appendix. The most important parameters are listed below:
\paragraph{Environment}
\begin{itemize}
    \item \textbf{State Space} Vector $\in \mathbb{R}^{27}$ generated from a single image, containing (i) entropy and BvsSB score of the IC model's prediction, (ii) IC model's layer-wise std-deviation, mean and norm, (iii) average F1-Score of the current IC model
    \item \textbf{Action Space} $a \in \{label, notLabel\}$
    \item Budget: 800
    \item Reward Shaping: False
\end{itemize}
\vspace{3mm}
\begin{minipage}{0.48\textwidth}
    \paragraph{Agent}
    \begin{itemize}
        \item Dense 24: LeakyReLU
        \item BatchNorm
        \item Dense 12: LeakyReLU
        \item BatchNorm
        \item \textbf{Policy:} Softmax greedy
    \end{itemize}
\end{minipage}
\begin{minipage}{0.48\textwidth}
    \paragraph{General}
    \begin{itemize}
        \item Number of interactions: 12000
        \item Exploration: 4000
        \item Conversion: 4000
        \item Evaluation runs: 15
    \end{itemize}
    \vspace{5.5mm}
\end{minipage}\\[5mm]
\newpage
The agent is trained with games of budget 800 for a total number of 12000 interactions. The exploration and conversion periods define the progression of the greed parameter $\tau$ for the softmax greedy policy (Eq. \ref{eq:softmaxGreedy}) of the agent. During the exploration period $\tau$ is 1, during the conversion period it is gradually lowered to the final value of 0.2. Every interaction thereafter stays at $\tau = 0.2$\\[1mm]
The training progress of the agent is visualized by the cumulated loss and reward per game. Each new game is marked with a vertical dotted line.
\begin{figure}[H]
    \centering
    \includegraphics[width=\textwidth]{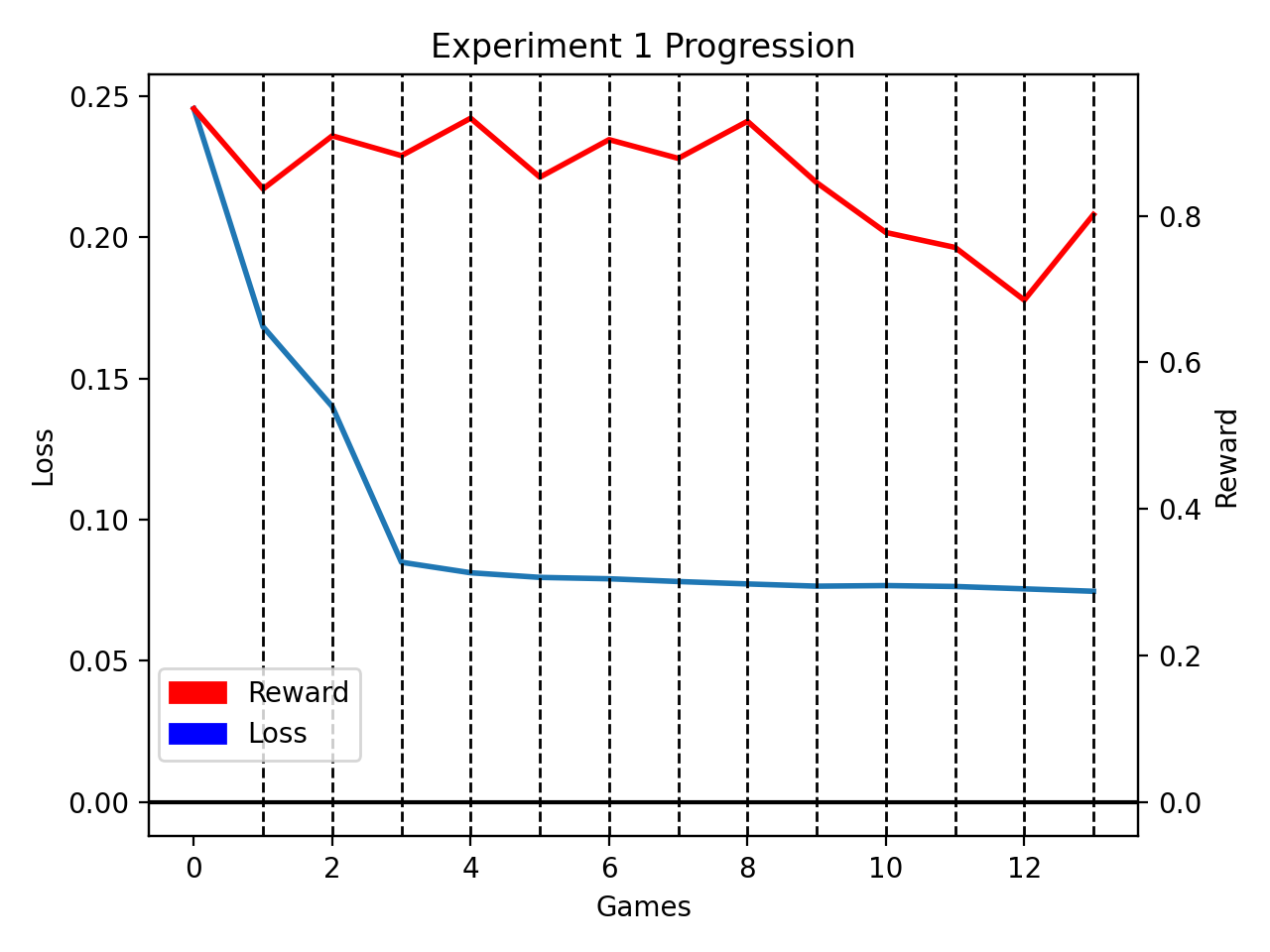}
    \caption{Experiment 1 - Loss and reward progression}
    \label{fig:exp1_prog}
\end{figure}
The script caches the top performing agent throughout the whole training process and uses that checkpoint for the evaluation, rather than the latest state of the agent. The evaluation is conducted as described in Section \ref{sec:alEval}. To consolidate the results, the evaluation is repeated 15 times and the curves are averaged over all runs.
\begin{figure}[H]
    \centering
    \includegraphics[width=.9\textwidth]{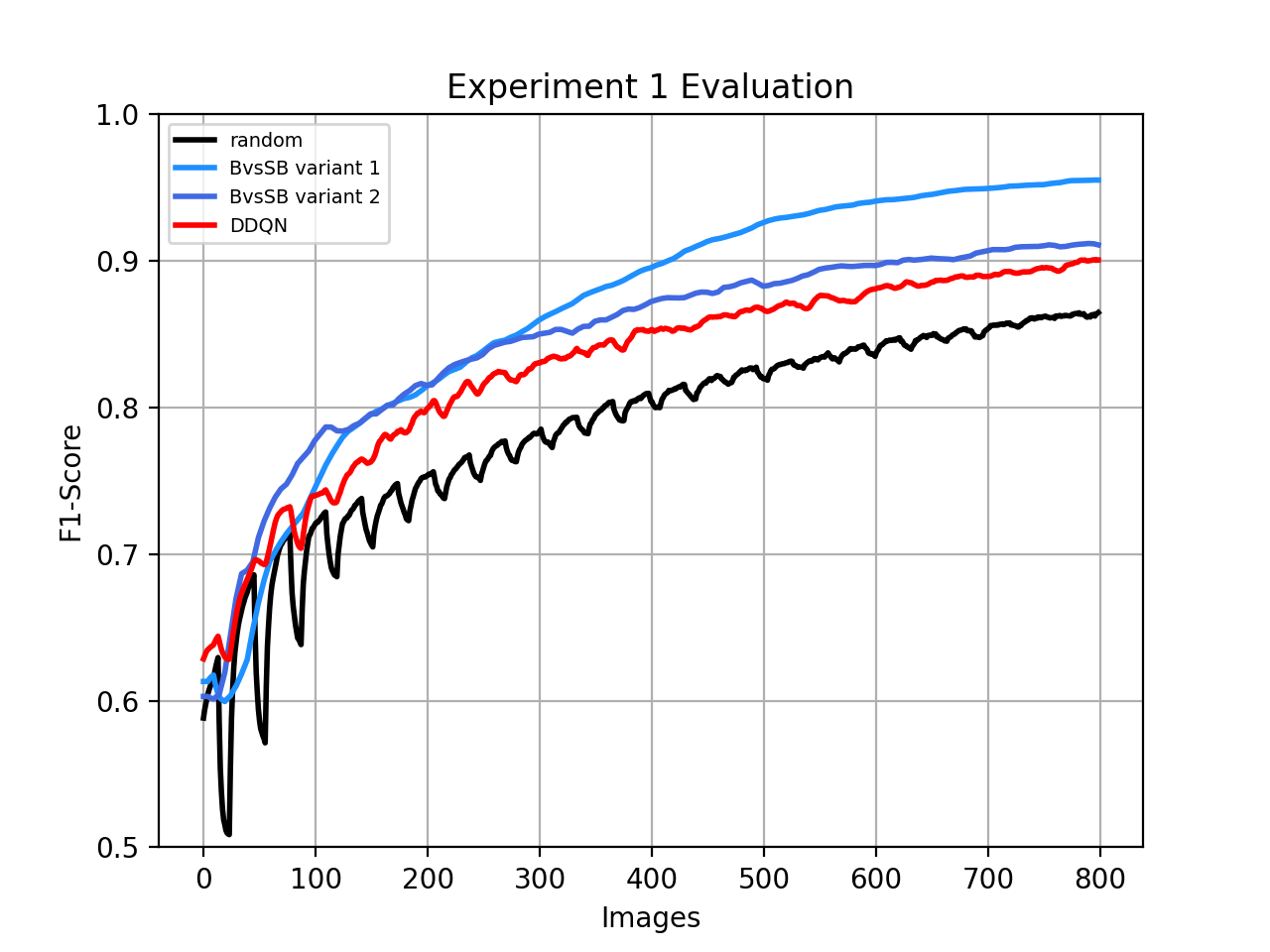}
    \caption{Experiment 1 - Evaluation of trained agent compared to the baselines}
    \label{fig:exp1_eval}
\end{figure}
As stated in Section \ref{sec:alEval}, a higher curve profile indicates a higher performance during evaluation. Even though a significant lift of the agent compared to the random baseline can be observed, it lacks behind the baselines, in the same manner. The following values of Table \ref{tab:exp1_eval} are averaged in a window of [-10, +10] around each evaluation point.
\begin{table}[H]
    \centering
    \caption{Experiment 1 - Comparison of the agent's performance at fixed points, represented by the F1-Score of the IC model}
    \begin{tabular}{c | c | c | c}
        & 100 & 400 & 800 \\
        \hline
        BvsSB 1 & 0.77 & 0.90 & 0.96 \\
        BvsSB 2 & 0.78 & 0.87 & 0.91 \\ 
        \textbf{DDQN} & 0.74 & 0.85 & 0.90  \\
        Random & 0.72 & 0.81 & 0.86
    \end{tabular}
    \label{tab:exp1_eval}
\end{table}
None of the baselines used the exact state representation (number of presented images (see Sec. \ref{sec:baselines})) and is directly comparable to the agent. However, both variants of the BvsSB sampling are computationally inexpensive and use a very limited set of information compared the agent and therefore should ideally be matched or succeeded by the agent. A detailed discussion of all trained agents and the employed baselines can be found in the conclusion.

\paragraph{Reward Shaping}
A second experiment with reward shaping was unsuccessful, due to the agent navigating himself into a seemingly locking state during evaluation. The training agent training did not show abnormal behavior, however, during evaluation the agent always reaches a point where it rejects every presented image and therefore cannot finish the evaluation game. This issue persisted even after many restarts of the training and ultimately was abandoned due to time constraints.

\section{Discussion: Miss-formulation of the RL environment}
The obtained results show the applicability of RL agents in an active learning setting. This however, was already shown by the baseline paper, whose workflow was implemented in this experiment. Even though the setting changed from Named Entity Recognition (NER) to image classification.\\
The main difference between the trained agent and the vastly superior BvsSB sampling methods is the contextualization of each decision. While the RL agent only is presented a single image and needs to decide wether to label the image or not, the baselines are presented the full dataset (BvsSB Variant 1) or a subset of the unlabeled data (BvsSB Variant 2). This difference is likely to be the main source of the lacking performance of the RL agent.\\
The approach of this first experiment inherits two issues: (i) The agent has no form of context for its decision. It can only decide, based on the presented state (based on one image) and its own internal modeling of the problem, which likely will be based on some sort of threshold mechanism. If that internal model is flawed at some point of the state space it can happen that the agent will refuse every image presented to it. This happened multiple times during training and was only overcome by restarting the whole training process. (ii) The states have a very weak dependency between each other. Every time the agent makes a decision the presented image is replaced by a random image from a large unlabeled dataset $|U| \approx 60000$. Since the impact of a single added image on the environment is minimal in most cases, this can be compared to a labyrinth problem, where the agent is randomly relocated after each move (the follow up state $s_{i+1}$ is very weakly dependent on the previous state $s_i$ and action $a_i$).

\chapter{Experiment 2}\label{chap:exp2}
\section{Experiment: Remodeling the Environment}
In order to improve the setup and overcome the issues discussed in the previous section, a new state and action space for the AL environment is proposed. \\
Instead of single images, a sample of size $s$ of multiple images is presented to the agent. The agent now decides if (i) one of these images should be added to the labeled dataset or (ii) no image should be added. In case (i) the chosen image is added to the labeled set and replaced by a random image from the unlabeled set, while the rest of the image sample stays the same. In case (ii) no image is added to the labeled set and a random image from the sample is replaced. \\
In addition to the previously discussed issues, this setup improves on the comparability of the training process and the standard AL workflow by moving closer to the intuition of AL that the sampling agent should be able to choose the most promising datapoint out of a sample. It is expected that this will increase the training efficiency and the predictive power of the agent.\\
This approach changes the action space of the environment to $s + 1$, where the additional action represents the \textit{add no image} action.

\newpage
\section{Execution and Results}
Again, a full specification of the setup can be found in the appendix. The most important parameters are listed below:
\paragraph{Environment}
\begin{itemize}
    \item \textbf{State Space} Vector $\in \mathbb{R}^{35}$ generated from a sample of images, containing (i) entropy and BvsSB score of the IC model's prediction for each image, (ii) IC model's layer-wise std-deviation, mean and norm, (iii) average F1-Score of the current IC model
    \item \textbf{Action Space} $a \in [0, 1,  ... \hspace{1mm} , s + 1]$
    \item Budget: 800
    \item Sample size $s$: 5
    \item Reward Shaping: False
\end{itemize}
\vspace{3mm}
\begin{minipage}{0.48\textwidth}
    \paragraph{Agent}
    \begin{itemize}
        \item Dense 48: LeakyReLU
        \item BatchNorm
        \item Dense 24: LeakyReLU
        \item BatchNorm
        \item \textbf{Policy:} Softmax greedy
    \end{itemize}
\end{minipage}
\begin{minipage}{0.48\textwidth}
    \paragraph{General}
    \begin{itemize}
        \item Number of interactions: 12000
        \item Exploration: 4000
        \item Conversion: 4000
        \item Evaluation runs: 15
    \end{itemize}
    \vspace{5.5mm}
\end{minipage}\\[5mm]
Analogously to experiment 1, the agent is trained for 12000 interactions, with 4000 interactions exploration and 4000 interaction conversion. The sample size $s$ for this experiment is set to 5, which results in an action space of $\{0,1,2,3,4,5\}$ with action $5$ being \textit{add no image}.\\
The agent was scaled up by a factor of 2 to compensate for the increased complexity of the state space.\\
Again, the training progress is visualized by the cumulated loss and reward per game.
\begin{figure}[H]
    \centering
    \includegraphics[width=.8\textwidth]{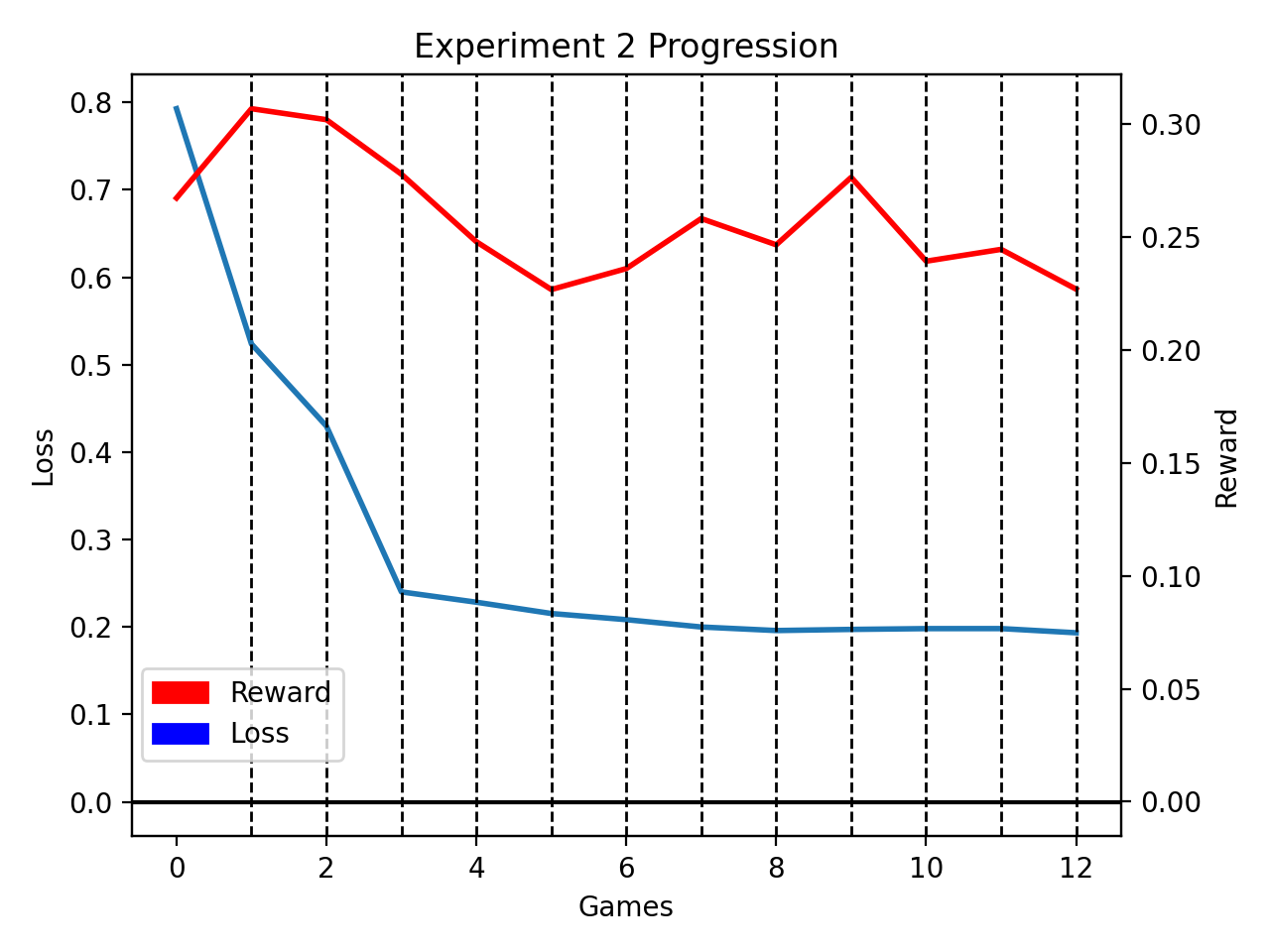}
    \caption{Experiment 2 - Loss and reward progression}
    \label{fig:exp2_prog}
\end{figure}
The obtained performance is measured over 15 evaluation runs and compared to the baselines.
\begin{figure}[H]
    \centering
    \includegraphics[width=.8\textwidth]{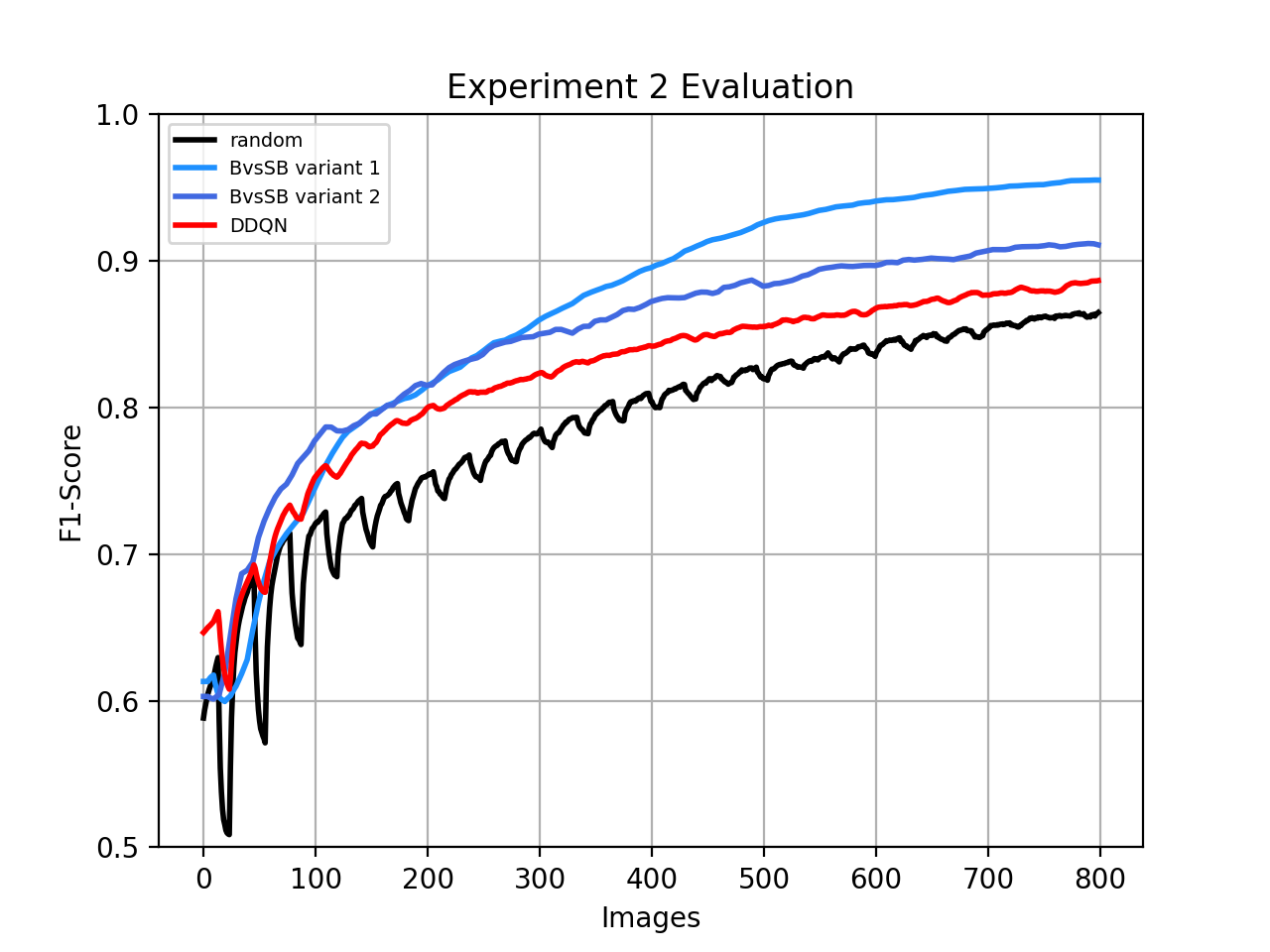}
    \caption{Experiment 2 - Evaluation of trained agent compared to the baselines}
    \label{fig:exp2_eval}
\end{figure}
The result is very comparable to the plot of experiment 1 (Fig. \ref{fig:exp1_eval}). To draw a comparison, the agents from both experiments are plotted.
\begin{figure}[H]
    \centering
    \includegraphics[width=.8\textwidth]{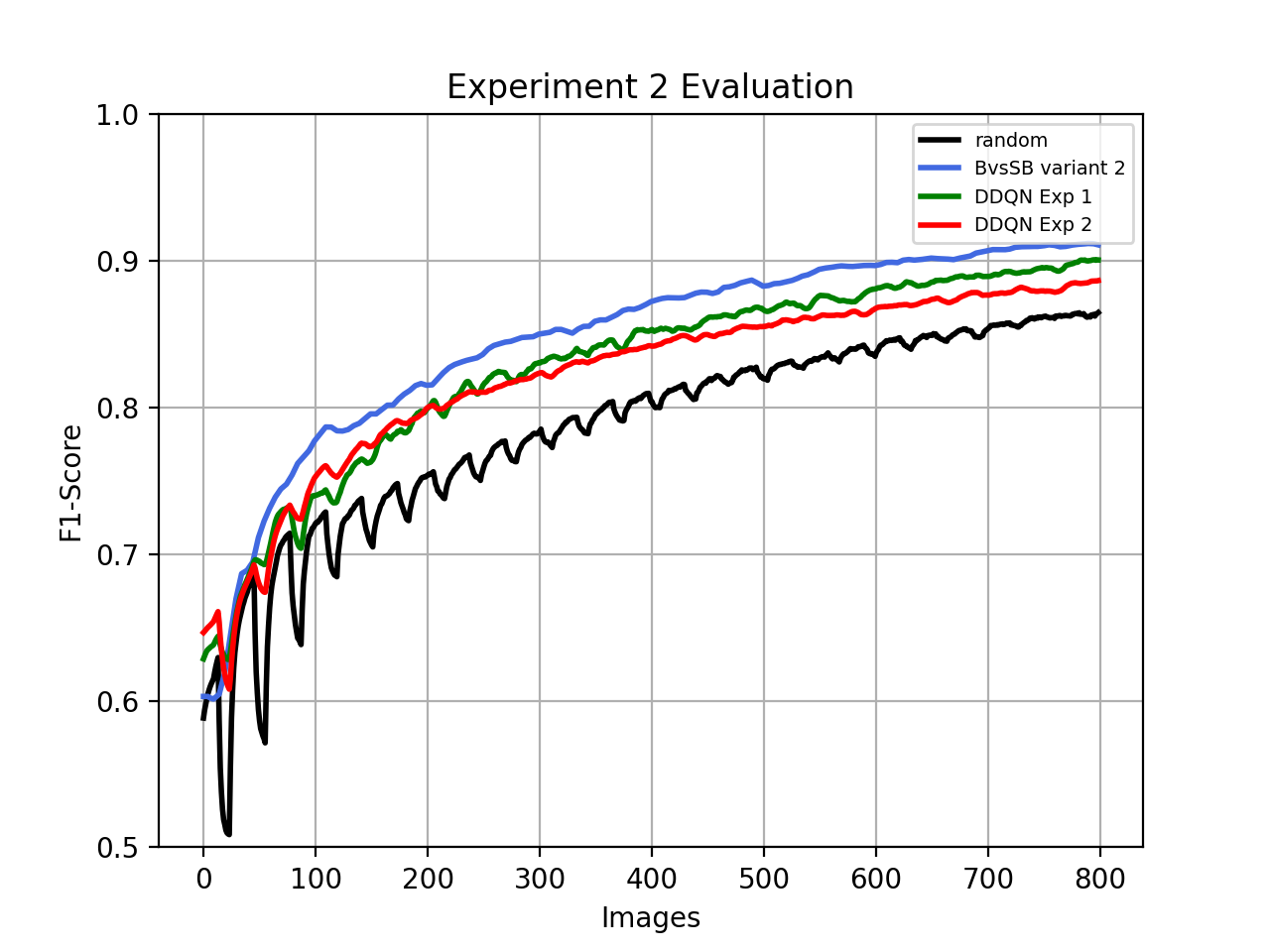}
    \caption{Comparison of the agents of experiment 1 and 2}
    \label{fig:exp2_eval2}
\end{figure}
From Figure \ref{fig:exp2_eval2}, it is apparent that the new agent actually performs worse than the agent from experiment 1. To confirm this the following table contains averaged measurements from different points of the evaluation.
\begin{table}[H]
    \centering
    \caption{Experiment 2 - Comparison of the agents' performances at fixed points, represented by the F1-Score of the IC model}
    \begin{tabular}{c | c | c | c}
        & 100 & 400 & 800 \\
        \hline
        BvsSB 2 & 0.78 & 0.87 & 0.91 \\ 
        DDQN Exp 1 & 0.742 & 0.853 & 0.900  \\
        DDQN Exp 2& 0.756 & 0.843 & 0.886  \\
        Random & 0.72 & 0.81 & 0.86
    \end{tabular}
    \label{tab:exp2_eval}
\end{table}
While for experiment 1 no baseline was directly comparable to the agent, for experiment 2, the agent uses the exact setup of baseline ``BvsSB Variant 2''. As stated in Section \ref{sec:baselines}, the only addition to the baseline is a decaying threshold that controls how many images are rejected on average.

\section{Discussion: Quality of the Reward signal}\label{sec:exp2_disc}
Considering the firm expectation of increased performance, this experiment needs to be marked as a failure. The updated state contains all information that the previous version did, and even adds context to each decision of the agent. The fact that agent 2 performs worse than agent 1 indicates a mistake in the implementation, or a misconception of the environment. \\
The implementation has been thoroughly checked and external libraries have been applied to the problem to verify the result. The comparison to the external libraries reinforced the results and did not reveal any new insights and consequently are omitted in this thesis. \\
The following discussion will not focus on the quality of the implementation, but on possible misconceptions of the problem.\\[1mm]
Now that the state space contains all necessary information to at least match the baseline variant 2, another possible angle of improvement is the reward signal currently issued by the environment, which remained unchanged between experiment 1 and 2.\\
Two possible issues can be discussed here. (i) The impact of each individual image that is added to the labeled set is minimal, and in some situations even negative. This means that each decision the agent makes only has a very small influence on the environment and therefore the reward signal.
(ii) Without reward shaping, the payoffs for the agent are incredibly long-term. With an active learning budget of 800 the agent is issued a reward at most every 800 interactions. This long-term nature combined with the minimal impact of each individual decision results in a high amount of noise that might negatively impact the agent's training procedure.\\
These two statements stem from a phenomenon that can be observed in the evaluation of both experiments so far (Fig. \ref{fig:exp1_prog} and Fig. \ref{fig:exp2_prog}). While the loss curves behave as expected and monotonically decrease, the cumulated rewards actually decrease over time as well. This is unexpected behavior for a reinforcement learning agent and might signal a problem with the reward signal of the environment. Generally an agent should archive increasing performances (increasing reward curve) by fitting the reward signal of the environment better (decreasing loss curve). Since this behavior is not observed, on the basis of current observations it even is inverse, the next experiment tries to improve the quality of the reward signal. \\ [1mm]
Apart from the theoretical discussion above, a simple sanity check has been employed to examine the learning behavior of the agent.
During training, the states are recorded in the memory buffer. From this buffer the states can be retrieved, concatenated and the mean $\mu$ and standard deviation $\sigma$ for each feature can be measured.
The values of $\mu$ and $\sigma$ can be used to draw a sample $s$ from the state space by sampling each feature $m$ uniformly from a two standard deviation range.
\begin{equation*}
    s_m \sim Uniform(\mu_m - 2\sigma_m, \mu_m + 2\sigma_m) \hspace{4mm} \forall m=1 ... M
\end{equation*}
For this evaluation, the impact of the BvsSB- and entropy-score for individual images on the respective Q-value is tested. Concretely, the correlation of a feature in the state space with the respective Q-Value is measured. \\
\begin{figure}[H]
    \centering
    \includegraphics[width=.9\textwidth]{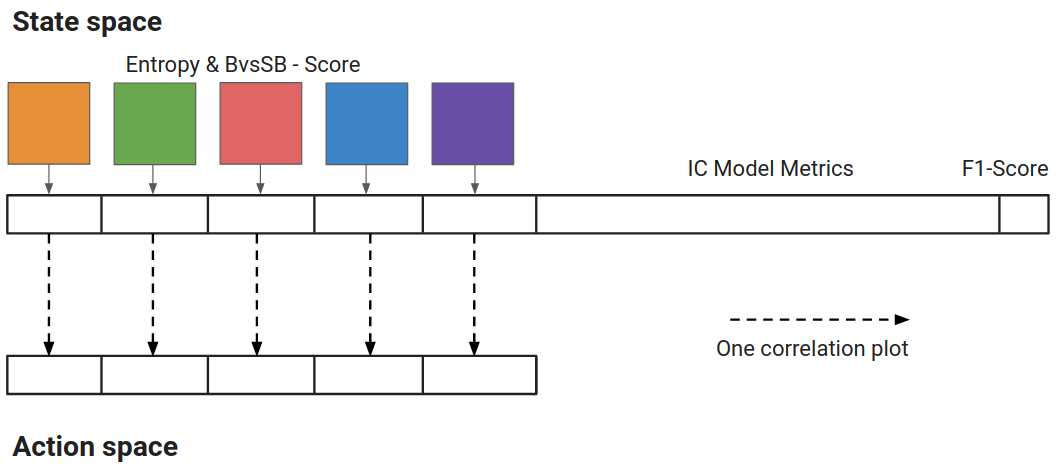}
    \caption{Visualization of the correlation plots}
    \label{fig:correlation}
\end{figure}
\begin{figure}[H]
    \centering
    \includegraphics[width=0.49\textwidth]{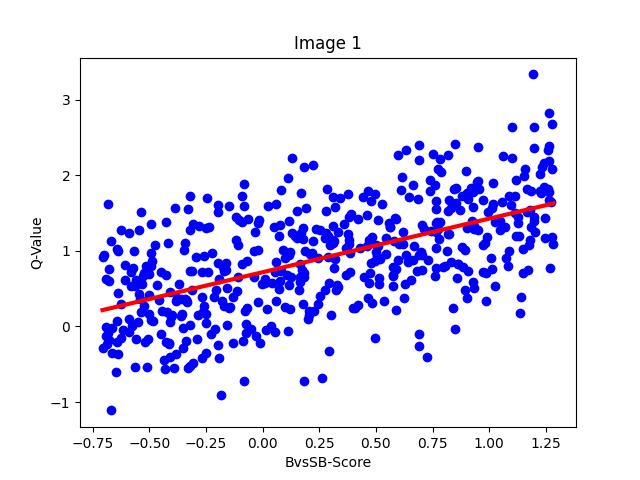}
    \includegraphics[width=0.49\textwidth]{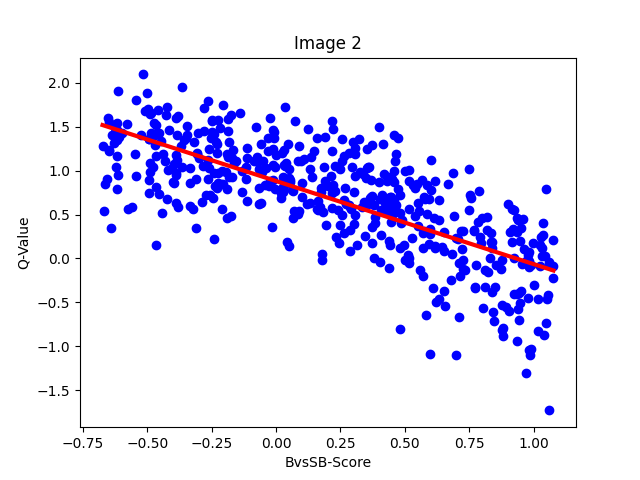}
    \includegraphics[width=0.49\textwidth]{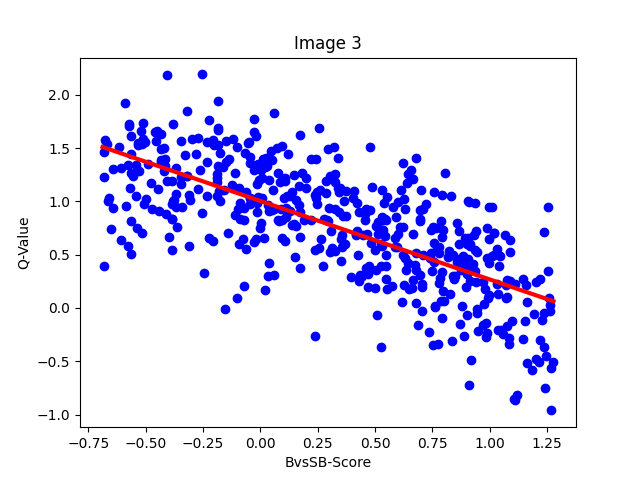}
    \includegraphics[width=0.49\textwidth]{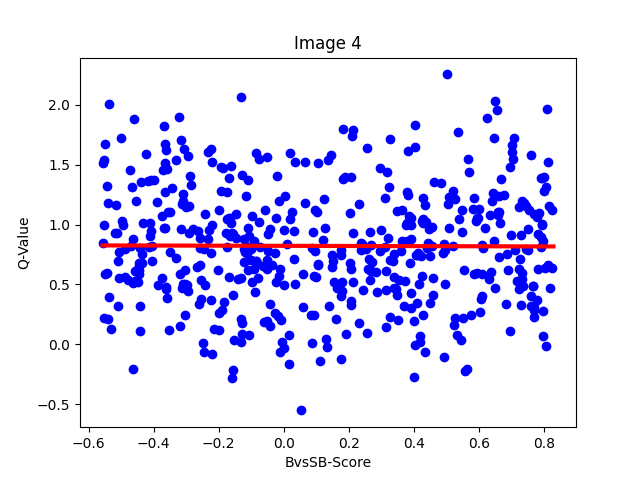}
    \includegraphics[width=0.49\textwidth]{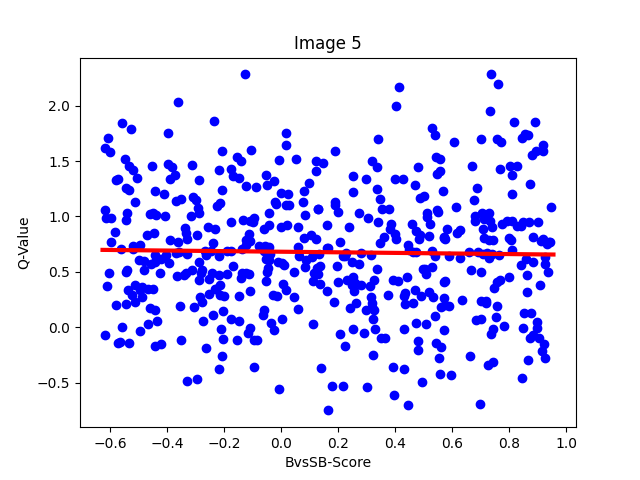}
    \caption{Experiment 2 - Impact of the BvsSB-score on the predicted Q-Value}
    \label{fig:BvsSB_2}
\end{figure}
\begin{figure}[H]
    \centering
    \includegraphics[width=0.49\textwidth]{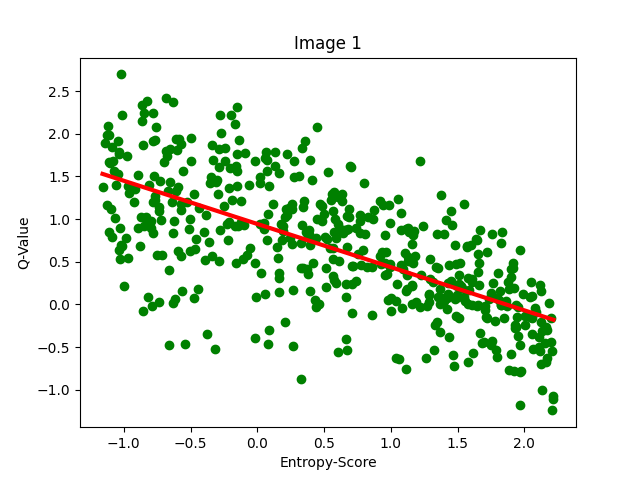}
    \includegraphics[width=0.49\textwidth]{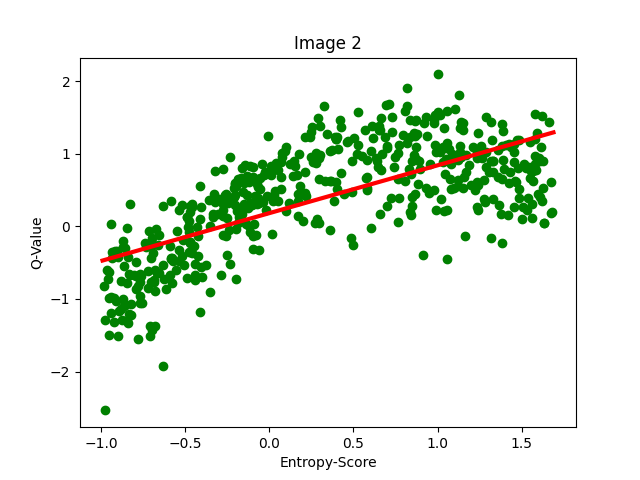}
    \includegraphics[width=0.49\textwidth]{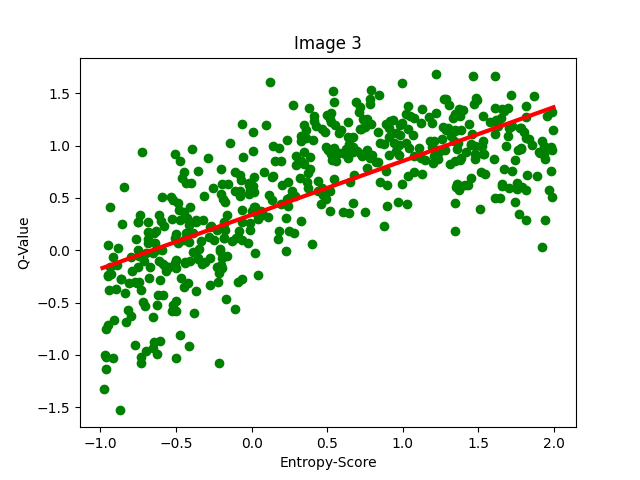}
    \includegraphics[width=0.49\textwidth]{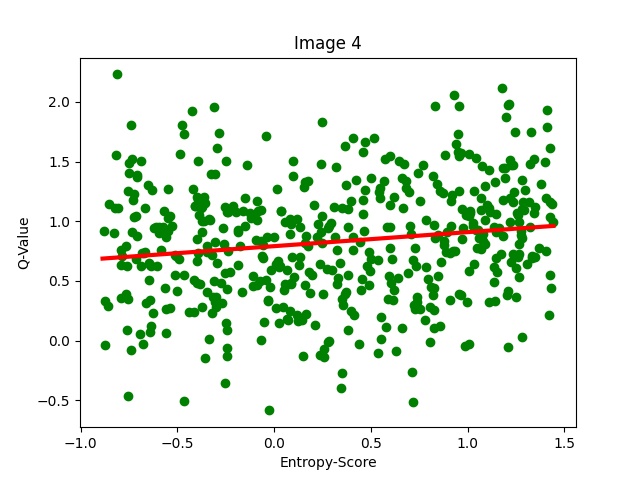}
    \includegraphics[width=0.49\textwidth]{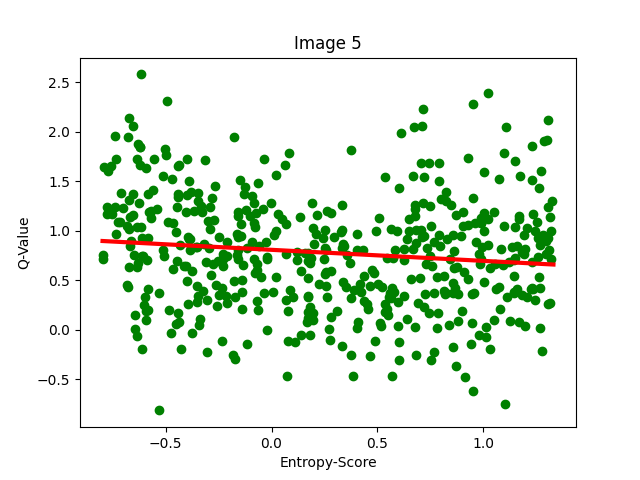}
    \caption{Experiment 2 - Impact of the entropy-score on the predicted Q-Value}
    \label{fig:entropy_2}
\end{figure}
The depicted behavior of the agent is a good indicator, that the agent is not capturing the relationship of BvsSB- and entropy-scores of an image with the potential value of that image for the AL process. Not only does the predicted Q-values fluctuate heavily, it rarely shows the expected positive trend that assigns high predicted Q-values to high BvsSB- or entropy-scores.

\chapter{Experiment 3}\label{chap:exp3}
\section{Experiment: Playing shorter Games}
To improve the quality of the reward signal, three changes to the environment are proposed. (i) Instead of considering a sample of single images for each state, the environment bundles multiple images and averages their metrics.
\begin{figure}[H]
    \centering
    \includegraphics[width=\textwidth]{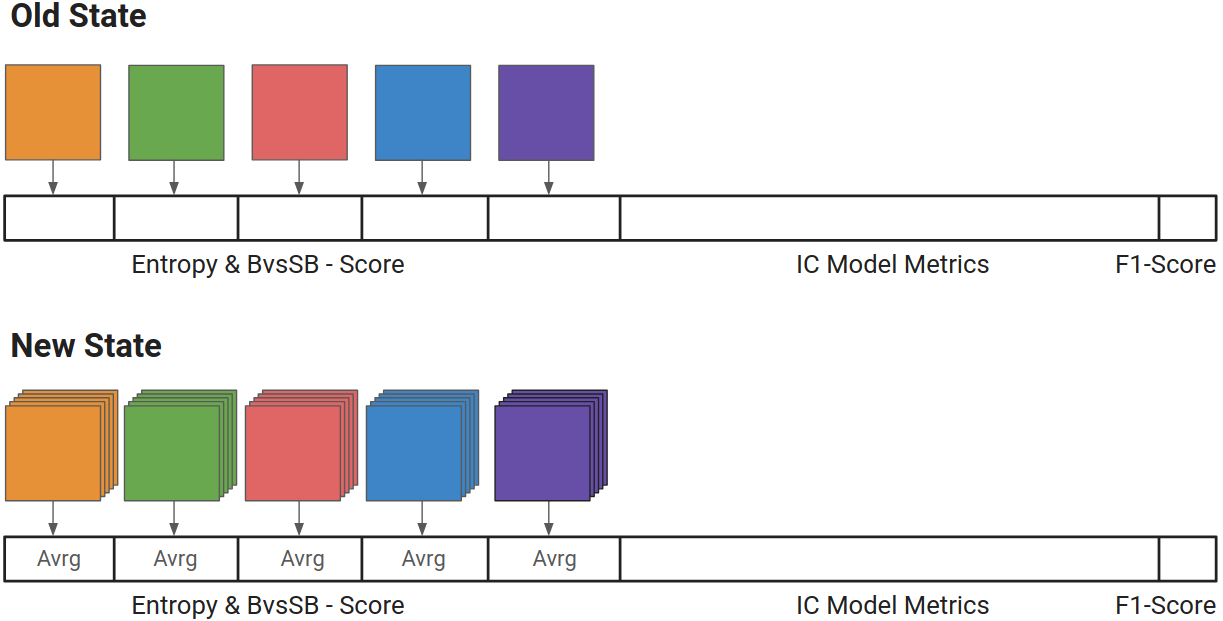}
    \caption{Visualization of the new state with bundled images instead of single images}
\end{figure}
Consequently, each interaction of the agent now adds multiple images to the labeled set instead of just one. This fixes the issue of minimal impact of the agent's decisions, discussed in the previous section.\\
(ii) Instead of playing one big game of active learning until the budget is exhausted and then fully reset the whole environment, play shorter sub-games with soft resets between them until the global budget is exhausted.
This setup uses no reward shaping, but a reward is issued to the agent after each sub-game. Sub-games are played until the environment reaches the global AL budget and fully resets.\\
To understand hard and soft resets the internal variables of the AL environment need to be considered.
\begin{itemize}
 \item {\color{blue} array}  \texttt{labeledDataset} - Labeled set for the active learning process
 \item {\color{blue} int} \texttt{budget} - Global active learning budget
 \item {\color{blue} int} \texttt{subgameLength} - Number of added images per sub-game
 \item {\color{blue} int} \texttt{addedImages} - Number of currently added images in the sub-game
 \item {\color{blue} float} \texttt{initalF1Score}  - F1-Score at the start of the sub-game
 \item {\color{blue} float} \texttt{currentF1Score}  - F1-Score after the last interaction
\end{itemize}
Consider the start of the training process for the agent. Initially the environment is hard-reset. The  \texttt{labeledDataset} is initialized with a seed set, the image classification model $\theta$ is trained and evaluated and \texttt{currentF1Score} and \texttt{initalF1Score} are set.
\begin{lstlisting}[escapeinside={(&}{&)}, caption={Hard Reset of the Environment in Experiment 3}, captionpos=t, label={alg:hardReset}]
def hard_reset():
    (&\texttt{labeledDataset}&) = initSeedSet()
    (&$\theta$&) = fit((&\texttt{labeledDataset}&))
    (&\texttt{initalF1Score}&) = (&\texttt{currentF1Score}&) = evaluate((&$\theta$&))
    (&\texttt{addedImages}&) = 0
\end{lstlisting}
At this point the agent starts to interact with the environment and adds images to the \texttt{labeledDataset}. The number of added images is tracked by \texttt{addedImages}. Once \texttt{addedImages} is equal or greater than \texttt{subgameLength} a reward is issued based on \texttt{currentF1Score - initalF1Score} and the environment is soft-reset. The soft updates \texttt{initalF1Score} and resets \texttt{addedImages}, but crucially keeps \texttt{labeledDataset} intact.
\begin{lstlisting}[escapeinside={(&}{&)}, caption={Soft Reset of the Environment in Experiment 3}, captionpos=t, label={alg:softReset}]
def soft_reset():
    (&\texttt{initalF1Score}&) = (&\texttt{currentF1Score}&)
    (&\texttt{addedImages}&) = 0
\end{lstlisting}
These sub-games are played until enough images are added to \texttt{labeledDataset} to exhaust the global \texttt{budget}. This is determined by $|$\texttt{labeledDataset}$|$ - $|$\texttt{seedSet}$|$ $>$ \texttt{budget}. Once that threshold is reached, the environment hard resets. This procedure is repeated until the desired number interactions for the agent training is reached.\\
To finish the enumeration, the third proposal is to (iii) introduce reward scaling by a constant factor. It has been observed, that the reward signal of the sub-games spans only a small area around 0. To provide a more pronounced signal to the agent, the reward should be scaled to ideally match the typical [0, 1] interval of reinforcement learning.

\section{Execution and Results}
\begin{figure}[H]
    \centering
    \includegraphics[width=\textwidth]{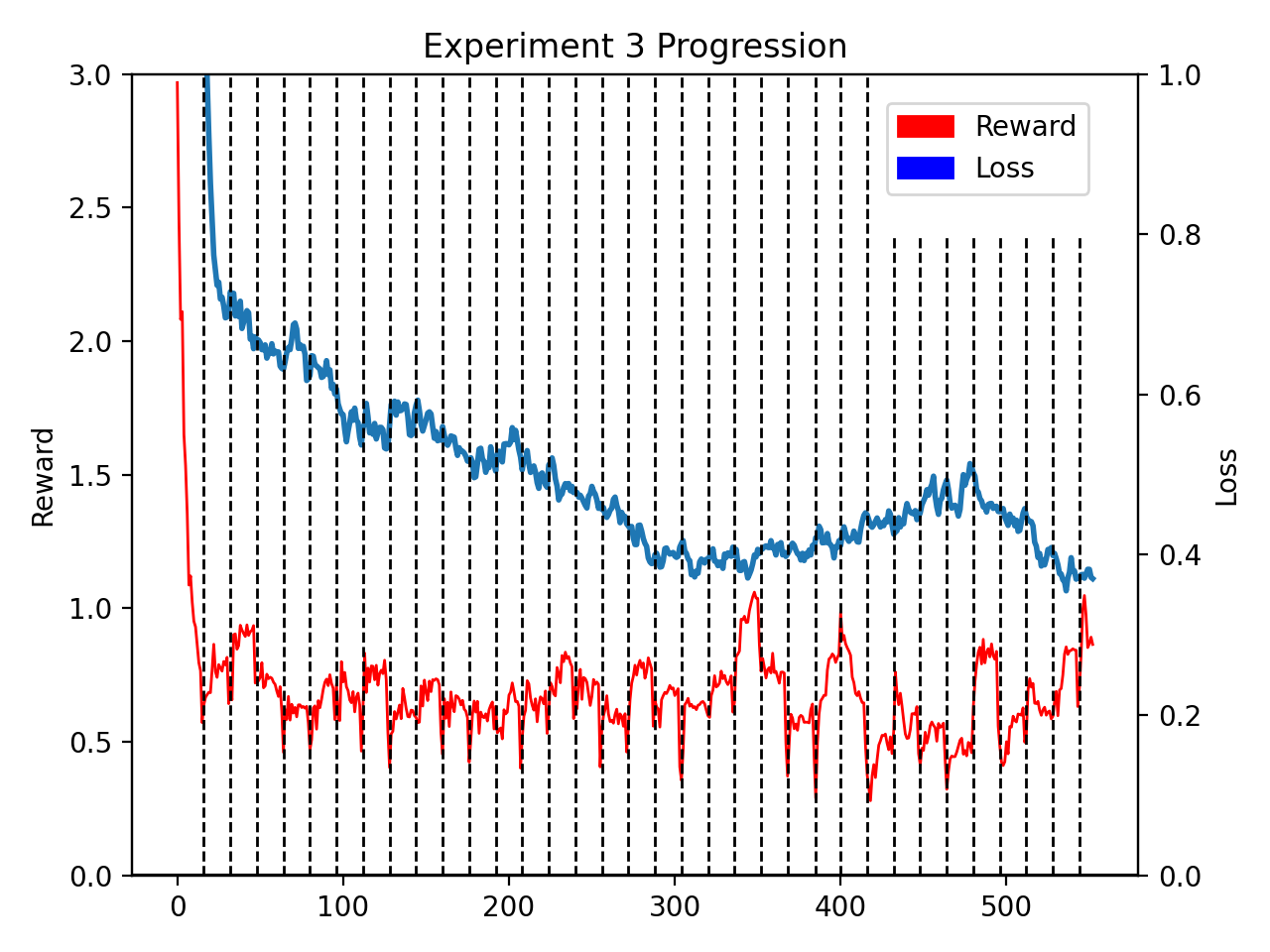}
    \caption{Experiment 3 - Loss and reward progression}
    \label{fig:exp3_prog}
\end{figure}

One last time, a full specification of the setup can be found in the appendix. The most important parameters are listed below:
\paragraph{Environment}
\begin{itemize}
    \item \textbf{State Space} Vector $\in \mathbb{R}^{35}$ generated from a sample of images, containing (i) averaged entropy and BvsSB scores of the IC model's prediction for each image bundle, (ii) IC model's layer-wise std-deviation, mean and norm, (iii) average F1-Score of the current IC model
    \item \textbf{Action Space} $a \in [0, 1,  ... \hspace{1mm} , s + 1]$
    \item Budget: 800
    \item Sample size $s$: 5
    \item Images to Bundle: 5
    \item Sub-Game Length: 50
    \item Reward Scaling: 40
\end{itemize}
\vspace{3mm}
\begin{minipage}{0.48\textwidth}
    \paragraph{Agent (Unchanged)}
    \begin{itemize}
        \item Dense 48: LeakyReLU
        \item BatchNorm
        \item Dense 24: LeakyReLU
        \item BatchNorm
        \item \textbf{Policy:} Softmax greedy
    \end{itemize}
\end{minipage}
\begin{minipage}{0.48\textwidth}
    \paragraph{General}
    \begin{itemize}
        \item Number of interactions: 8000
        \item Exploration: 3000
        \item Conversion: 3000
        \item Evaluation runs: 15
    \end{itemize}
    \vspace{5.5mm}
\end{minipage}\\[5mm]
For this experiment the agent is only trained for only 8000 interactions with 3000 interactions each exploration and conversion. Since 5 images are added per interaction, the AL budget is reached faster and more games are played. This leads to a faster conversion. The sample size $s$ for this experiment remains at 5.\\
A constant scaling of factor 40 was applied to the reward signal.\\
Each dotted vertical line in Figure \ref{fig:exp3_prog} still corresponds to a full game of active learning, each consisting of several sub-games (Hence there are multiple datapoints between each line).
\newpage
The evaluation procedure again, did not change compared to the previous experiments.
\begin{figure}[H]
    \centering
    \includegraphics[width=\textwidth]{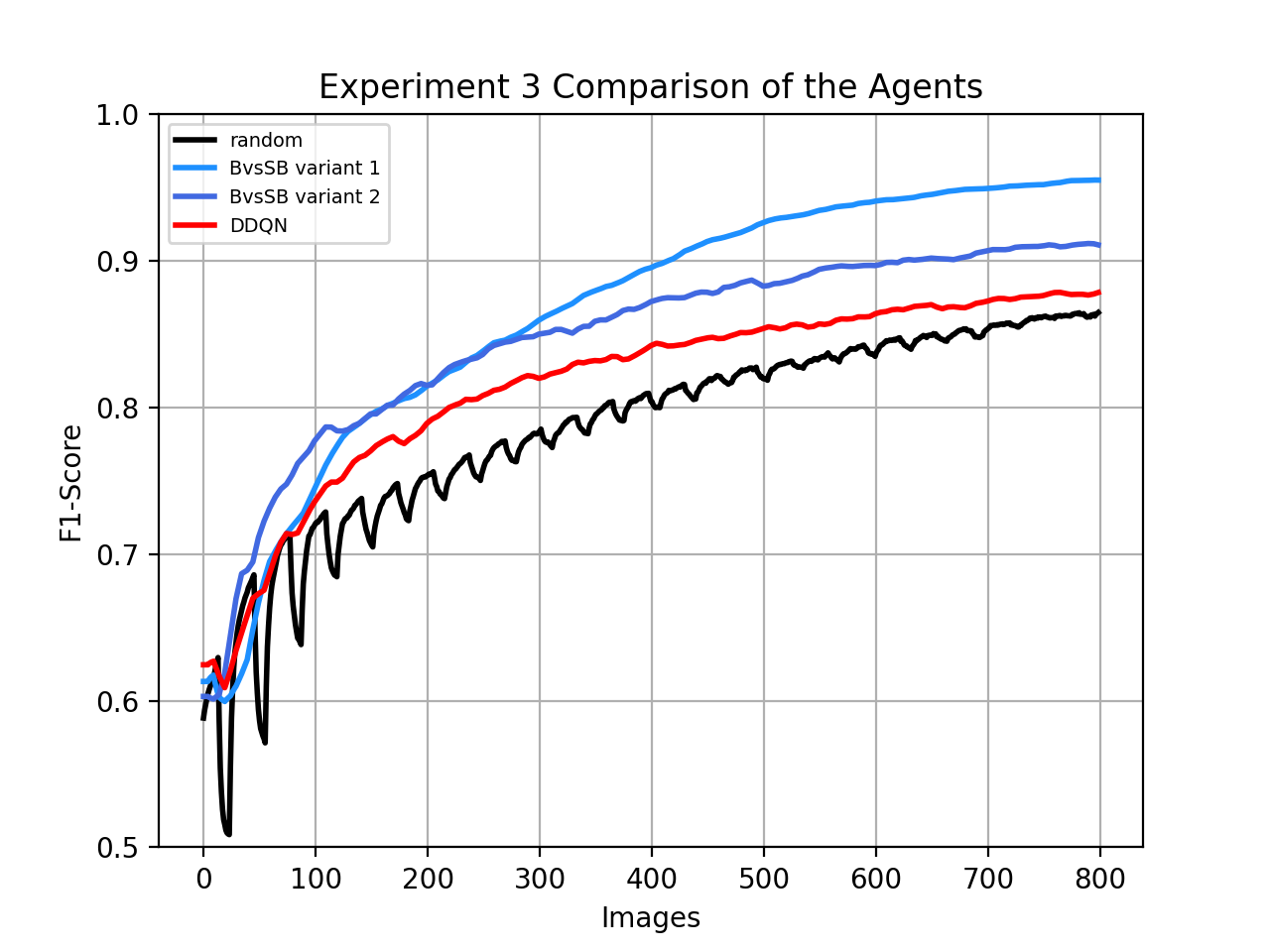}
    \caption{Experiment 3 - Evaluation of trained agent compared to the baselines}
    \label{fig:exp3_eval}
\end{figure}
As with experiment 2, the performance curve is very similar to experiments 1 and 2. To draw a comparison, all three agents are plotted next to each other and the their values are compared at the usual points in Table \ref{tab:exp3_eval}
\begin{figure}[H]
    \centering
    \includegraphics[width=\textwidth]{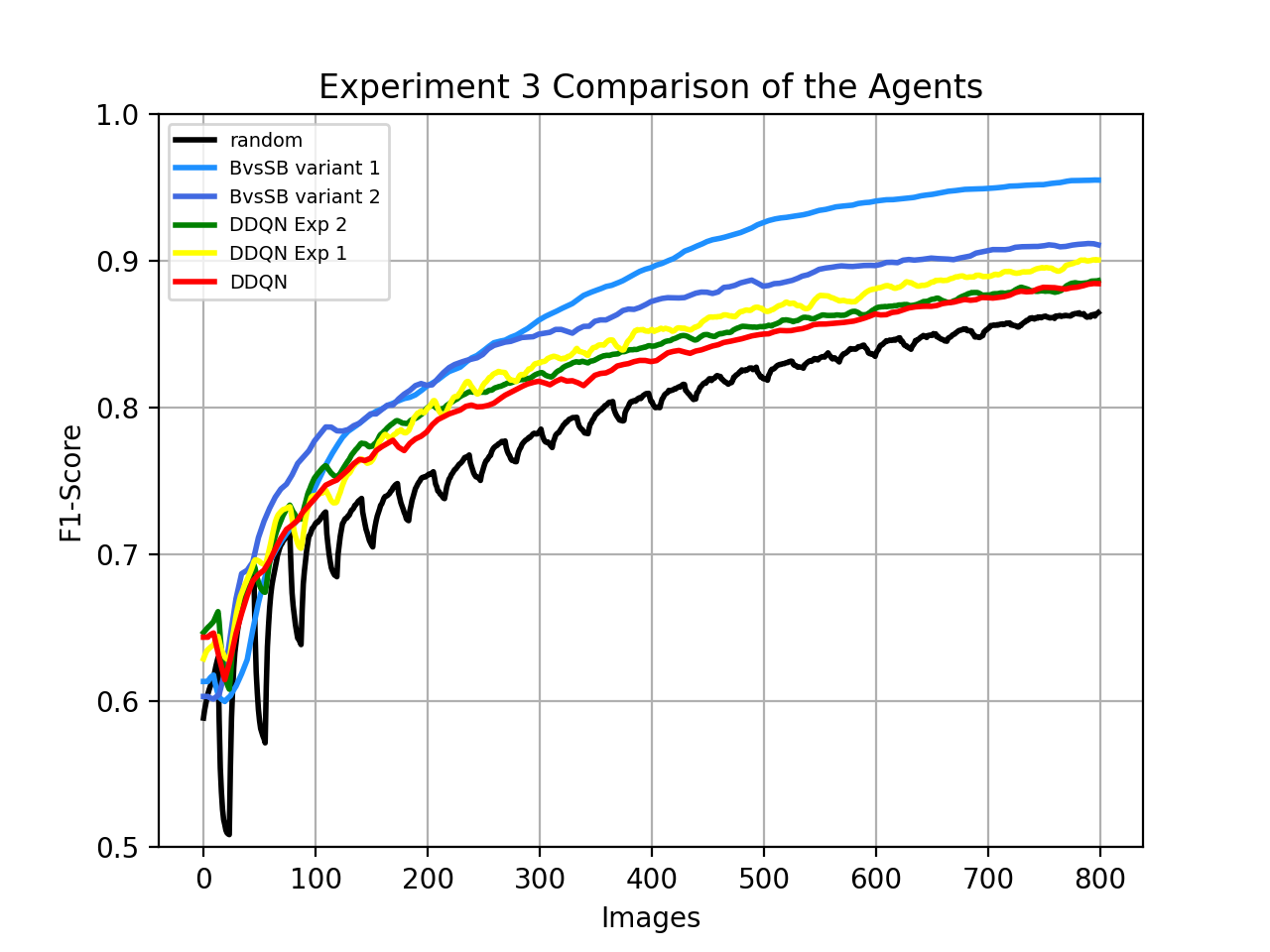}
    \caption{Experiment 3 - Evaluation of trained agents of all experiments}
    \label{fig:exp3_eval2}
\end{figure}
\begin{table}[H]
    \centering
    \caption{Experiment 3 - Comparison of the agents' performances at fixed points, represented by the F1-Score of the IC model}
    \begin{tabular}{c | c | c | c}
        & 100 & 400 & 800 \\
        \hline
        BvsSB 2 & 0.78 & 0.87 & 0.91 \\ 
        DDQN Exp 1 & 0.742 & 0.853 & 0.900  \\
        DDQN Exp 2 & 0.756 & 0.843 & 0.886  \\
        DDQN Exp 3 & 0.742 & 0.833 & 0.884  \\
        Random & 0.72 & 0.81 & 0.86
    \end{tabular}
    \label{tab:exp3_eval}
\end{table}

\newpage
\section{Discussion: Assessing the changes}
\begin{figure}[H]
    \centering
    \includegraphics[width=.65\textwidth]{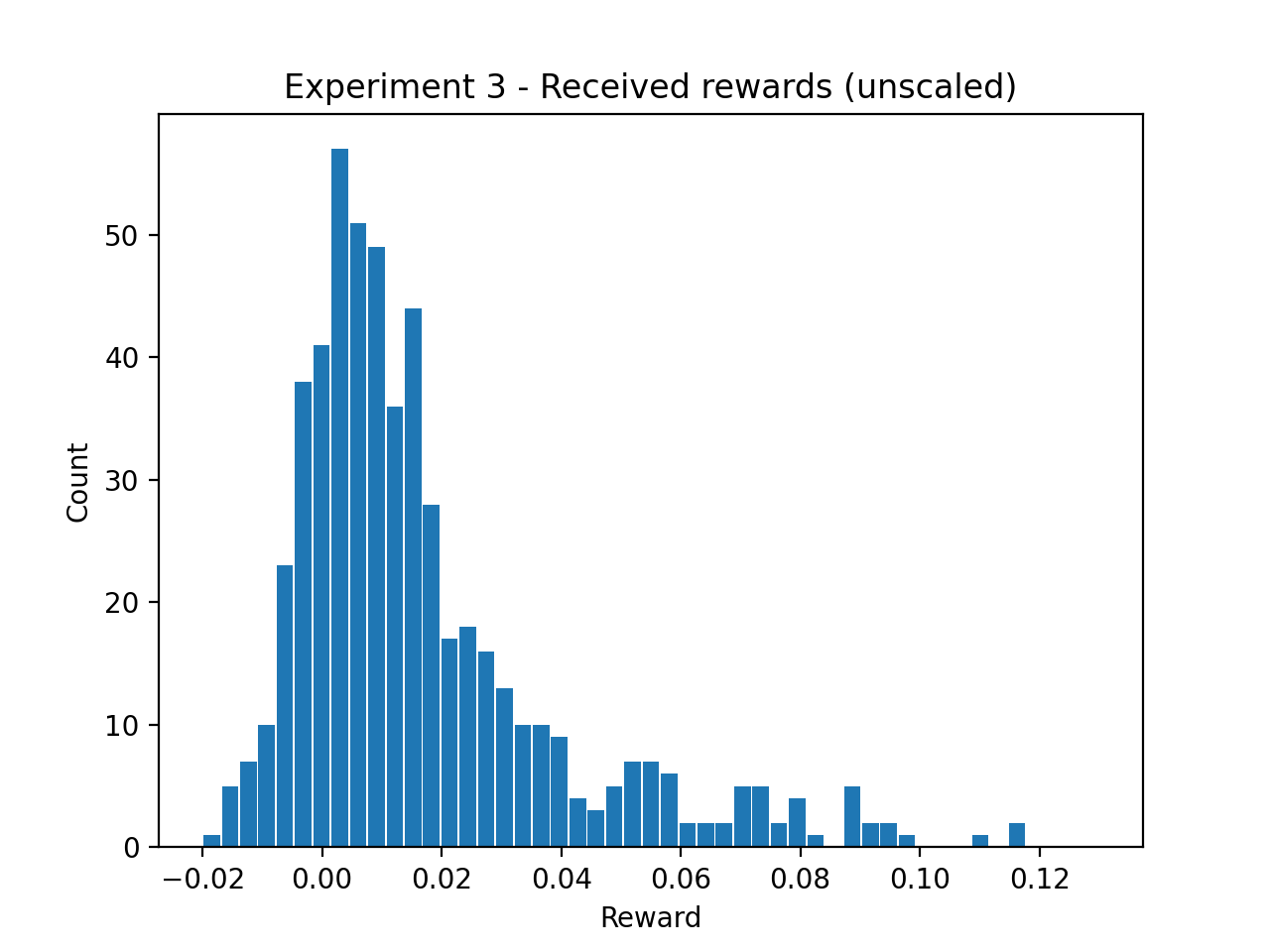}
    \centering
    \includegraphics[width=.65\textwidth]{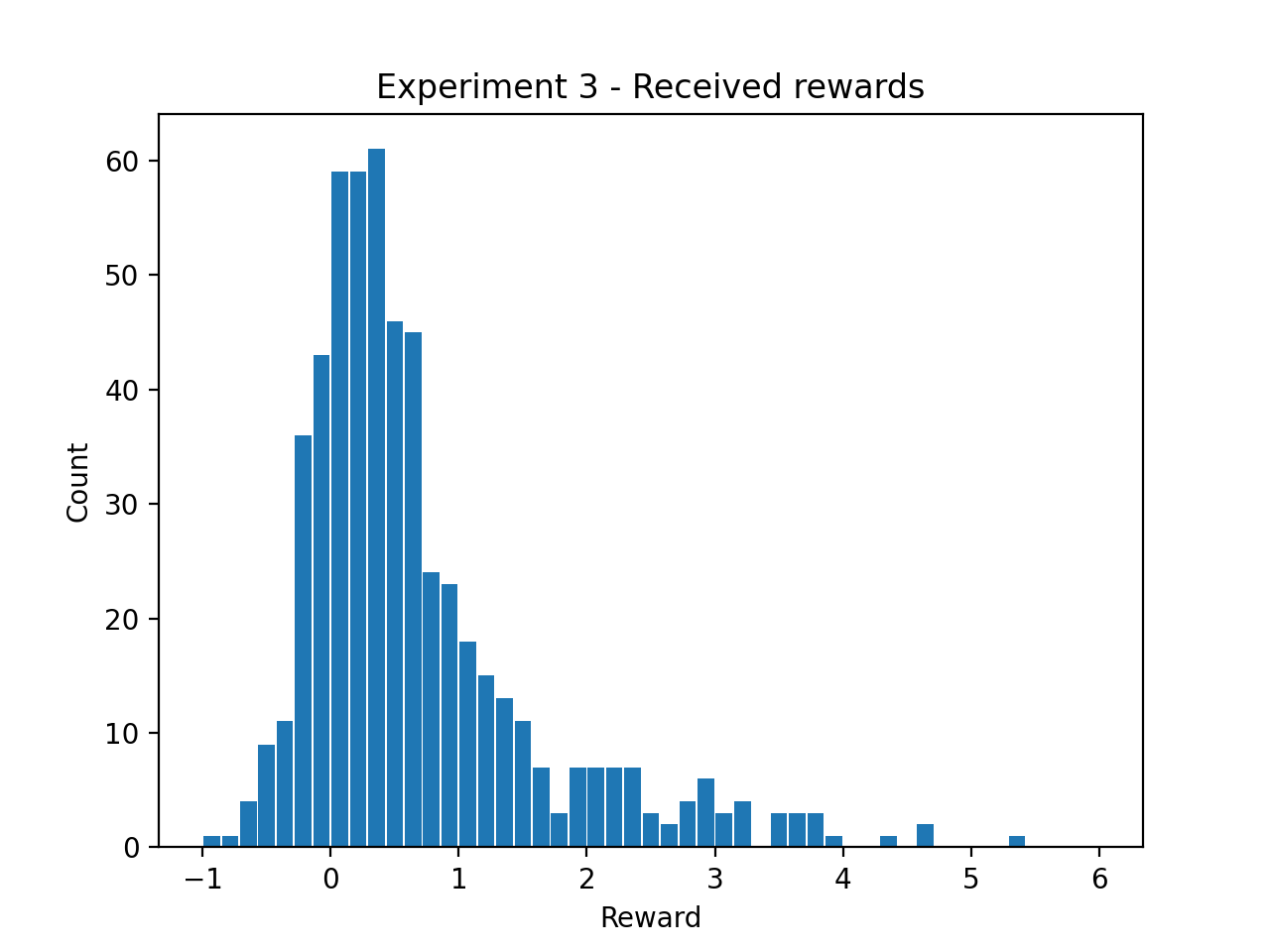}
    \caption{Experiment 3 - Histogram of the received rewards, unscaled (top) and  scaled by a factor of 40 (bottom)}
    \label{fig:exp3_hist}
\end{figure}

A decreased performance (Table \ref{tab:exp3_eval}), not only compared to experiment 1, but even to the direct predecessor, is an unexpected result. To tackle this issue each of the three proposed changes to the environment is discussed in isolation.
\paragraph{Reward Scaling} 
Since the distribution of the reward signal is difficult to read of Figure \ref{fig:exp3_prog} a histogram of the scaled and unscaled rewards was generated (Figure \ref{fig:exp3_hist}).
It is evident that the unscaled reward signal only provides minimal feedback for the agent's updates and therefore should be considered sub-optimal.\\
Considering the negative results from this experiment it cannot be stated that the scaled rewards ultimately provide better feedback, but the scaled distribution at least resembles a common distribution of RL rewards. Obviously, an amplification of the reward signal is of little use if the signal itself is flawed, or does not provide useful information.

\paragraph{Shorter Sub-Games}
To evaluate the impact of sub-games being played the received rewards during agent training (Fig. \ref{fig:exp3_prog}) are discussed. \\
Since, a lot more games are played compared to experiment 1 and 2 a lot more rewards are issued and a more detailed curve of the received rewards can be plotted.
\begin{figure}[H]
    \centering
    \includegraphics[width=.9\textwidth]{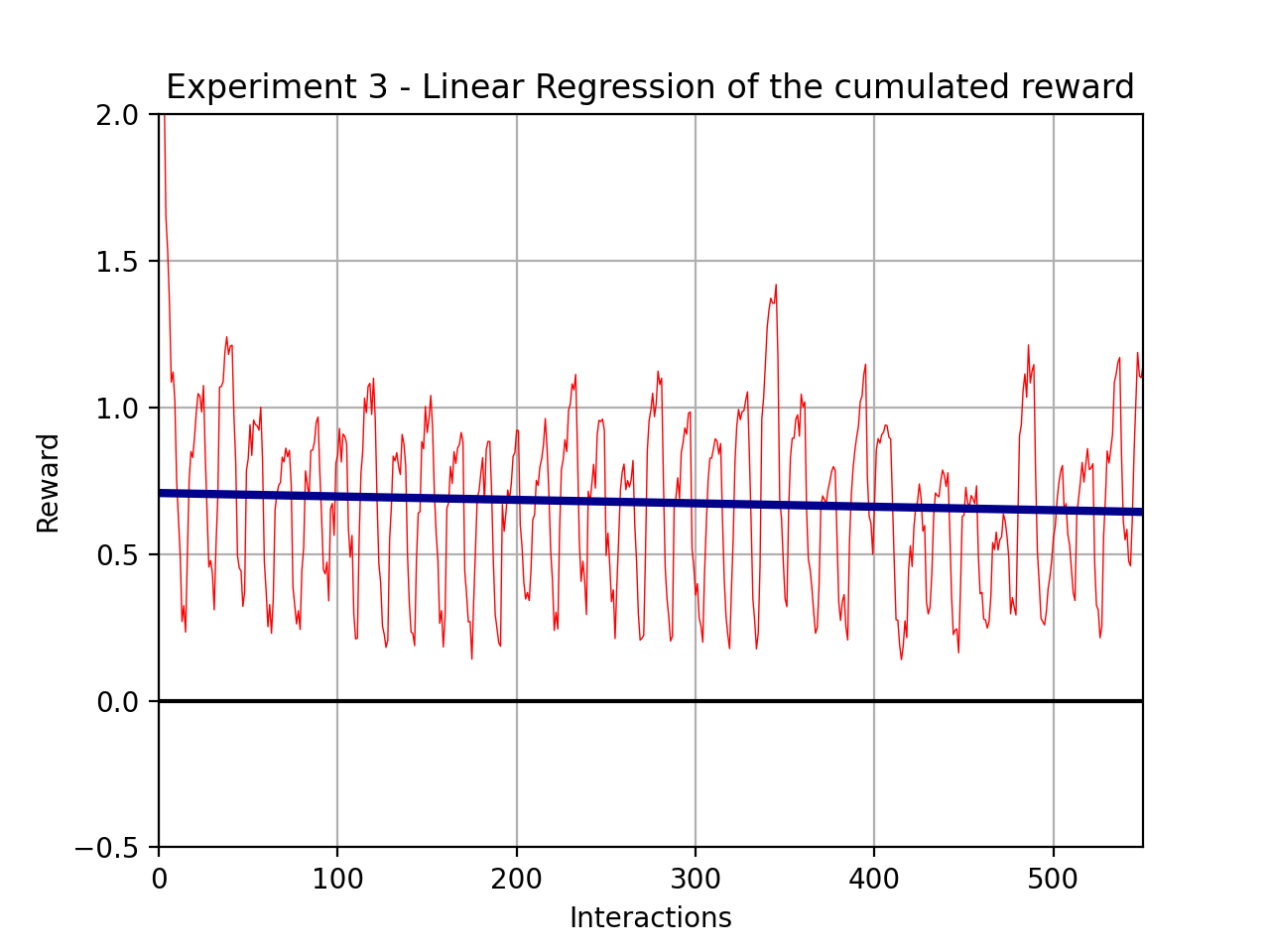} 
    \caption{Experiment 3 - Linear Regression of the received rewards}
    \label{fig:exp3_lr}
\end{figure}
The first observation is that the signal is heavily fluctuating. This is because the reward is generated by the improvement in F1-Score of the image classification model. Every time the environment is hard-reset (Alg. \ref{alg:hardReset}) the labeled set $L$ is small and every added bundle of images has a noticeable effect (producing positive rewards), while subsequent images have lower impact on the growing set $L$, and therefore produce lower, or negative rewards. Each peak of the reward signal in Figure \ref{fig:exp3_lr} corresponds to one hard reset of the environment.\\
The second observation is that the linear trend of the reward signal seems to be negative. This is a strong indicator that the agent is not learning a useful strategy for this problem. An analysis of the agents behavior can be found in paragraph ``General Analysis''.

\paragraph{Image Bundling}
Since the bundled images have not been tested in isolation it is not possible to evaluate on the specific impact of that update. Considering the negative outcome of this experiment, it can only be assumed that the positive impact was negligible or it did actually contribute negatively to the reward signal.

\begin{figure}[H]
    \centering
    \includegraphics[width=0.49\textwidth]{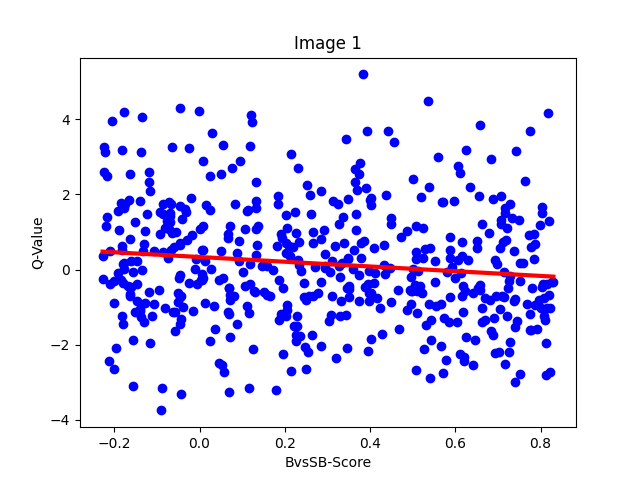}
    \includegraphics[width=0.49\textwidth]{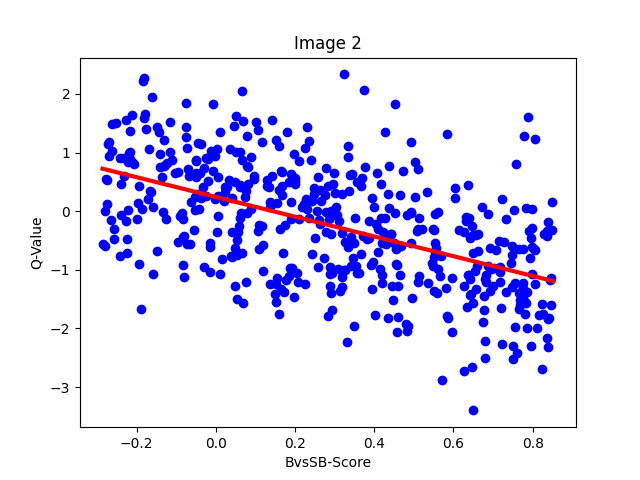}
    \includegraphics[width=0.49\textwidth]{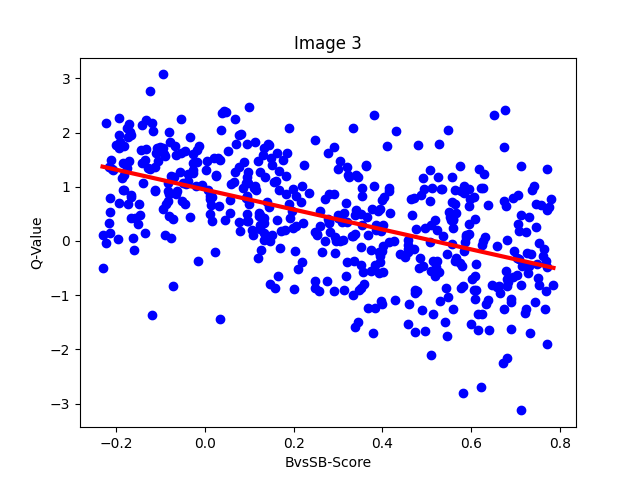}
    \includegraphics[width=0.49\textwidth]{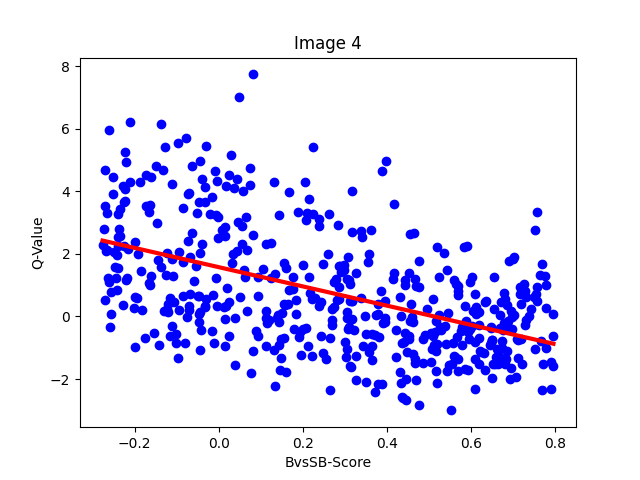}
    \includegraphics[width=0.49\textwidth]{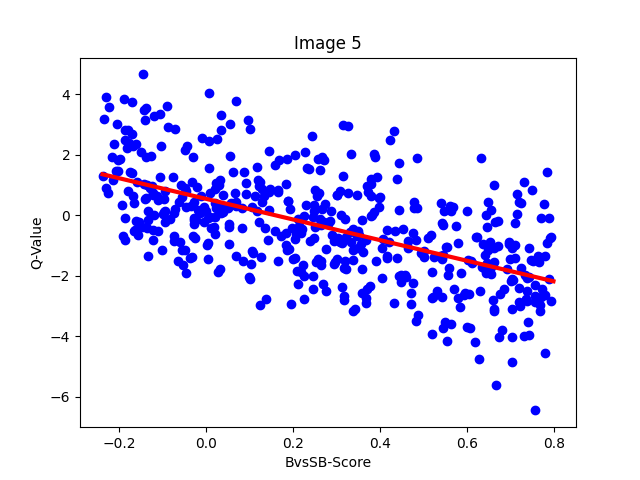}
    \caption{Experiment 3 - Impact of the BvsSB-score on the predicted Q-Value}
    \label{fig:BvsSB_3}
\end{figure}

\begin{figure}[H]
    \centering
    \includegraphics[width=0.49\textwidth]{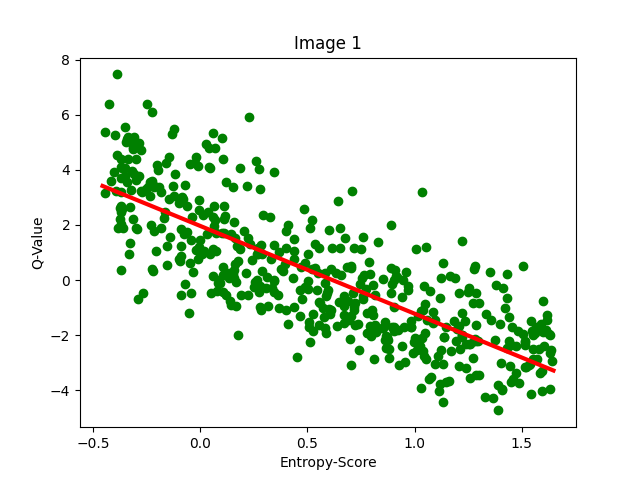}
    \includegraphics[width=0.49\textwidth]{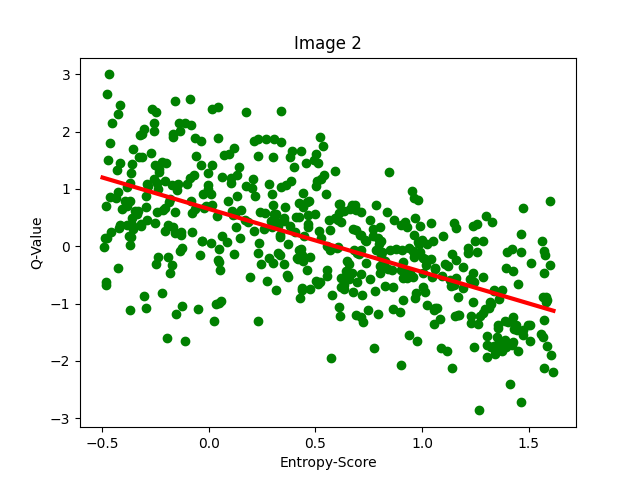}
    \includegraphics[width=0.49\textwidth]{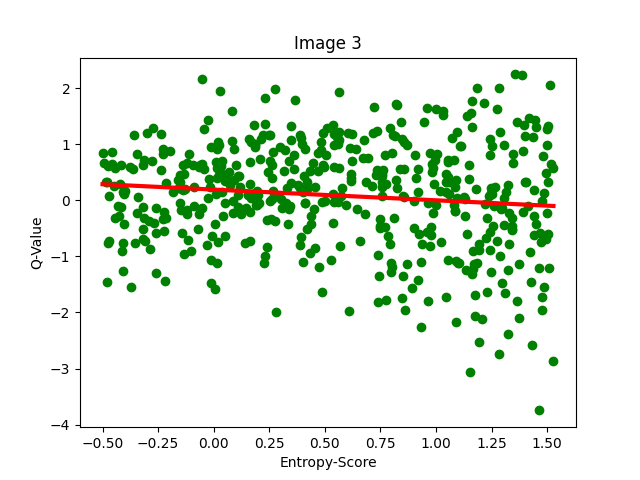}
    \includegraphics[width=0.49\textwidth]{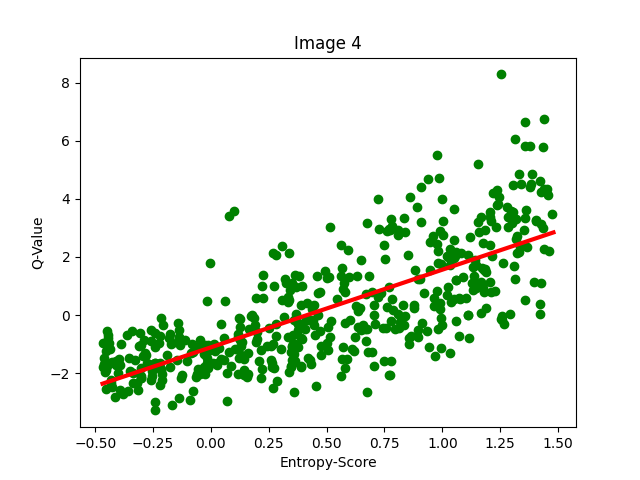}
    \includegraphics[width=0.49\textwidth]{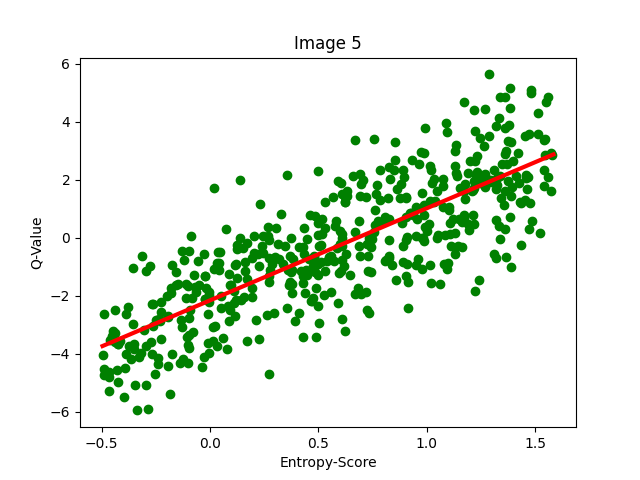}
    \caption{Experiment 3 - Impact of the entropy-score on the predicted Q-Value}
    \label{fig:entropy_3}
\end{figure}

\paragraph{General Analysis}
In this paragraph an analysis of the agent behavior is performed. To evaluate the learned policy, the impact of certain input feature on the predicted Q-Values are measured. For this, the process of drawing samples from a custom distribution, described in section ``Experiment 2 - Discussion'' (\ref{sec:exp2_disc}), is employed again (Fig. \ref{fig:BvsSB_3}, \ref{fig:entropy_3}). Similarly, the entropy- and BvsSB-Score are considered to be the most important features for the expected value of an image (and therefore it's predicted Q-Value).
The evaluation shows the same behavior from experiment 2 (Fig. \ref{fig:BvsSB_2} and \ref{fig:entropy_2}). 
As indicated in Section \ref{sec:exp2_disc} already, the overall behavior of the agent is unexpected to say the least. \\
The problems include: 
(i) Each of the 5 plots should depict the same behavior, since each plot corresponds to the same feature and simply uses a different bundle of images. 
(ii) The correlation of BvsSB-/entropy-score can be considered to be strictly positive, as both baselines for this work solely depend on these scores and reliably outperform any RL approach tested so far. Therefore the learned behavior of Figure \ref{fig:BvsSB_3} and \ref{fig:entropy_3} does not capture this correlation correctly.

\chapter{Conclusion}\label{chap:conclusion}
\section{Overall Results of the Experiments}
This conclusion, as for the thesis itself, will start at Fang's et. al. paper ``Learning how to Active Learn [...]'' \cite{howToActiveLearn}, which served as baseline paper. The authors of said paper are able to use RL to learn an active learning sampling strategy that consistently outperforms their baselines for a Named Entity Recognition (NER) task. While those results indicated the applicability of RL for active learning, a couple key differences to this thesis need to be pointed out. 
Fang et. al. used CRFs to classify their data, while this thesis uses CNNs. 
This introduces significant differences in the amount of needed training data for the model (200 sampled points in \cite{howToActiveLearn} vs. 800 in Experiment 1-3).
The different classification task for this thesis introduced the BvsSB-baseline that proved to be more powerful than entropy based sampling (as discussed in Section \ref{sec:baselines}), it remains untested if BvsSB would also outperform the baseline of \cite{howToActiveLearn} in their setting. \\
These differences were initially thought of as minor hurdles on the path of reproducing the results from \cite{howToActiveLearn}. Even though this thesis did not find a definitive cause for the lacking performance of experiment 1, they might be included in the underlying problems of that approach. \\
Apart from the trivial conclusion that this work did not match the expectations, the rest of this section will focus on the individual experiments. \\[2mm]
After replicating the algorithm from \cite{howToActiveLearn} and observing the lacking performance displayed in Figure \ref{fig:exp1_eval} the hypothesis was proposed that a poor formulation of the active learning problem in the reinforcement learning framework was the root cause.\\
The following experiments 2 and 3 were meant to prove this hypothesis and improve on the outlined problems. 
Experiment 2 implemented a more natural version of AL by presenting a sample of datapoints to the agent instead of single points, but failed to yield an improvement of performance.
The discussion in Section \ref{sec:exp2_disc} identified the noisy reward signal as a possible problem. 
The final experiment proposed several changes to the environment to improve the reward signal, which did not help the performance of the third approach,  but even lowered it  slightly below the previous experiments.\\
Even though many different improvements to the setup could still be implemented (see Chapter \ref{sec:futureWork}) the original hypothesis needs to be rejected. \\
Experiment 2 and 3 indicate that a reformulation of the AL problem does not lift the performance of the system on it's own. \\
However, the conducted evaluation of the impact of the entropy- and BvsSB-scores on the respective Q-Values (Figure \ref{fig:BvsSB_2}, \ref{fig:entropy_2}, \ref{fig:BvsSB_3} and \ref{fig:entropy_3}) indicate a sub-optimal learning process during agent training.
It is possible that by improving the training process the agent would be able to accurately capture the expected correlation and most likely improve it's performance during evaluation. \\[2mm]
At this point it is unclear if an improvement based on an other fundamental misconception would fix the poor performance of this system, or if it is a case of the ``right mix'' of techniques that need to be employed to result in a strong RL agent. To evaluate on this question, the outlined problems and the advanced techniques described in Chapter \ref{sec:futureWork} can be addressed. If one of the techniques mentioned of paragraph ``Advanced Concepts'' does indeed bring a significant lift in performance, it is very likely that the root problem was not the poor formulation of the RL environment, but instead the small size of labeled set $L$ (in case of advanced concept (iii) \cite{rwAL4}). On the other hand, if a comprehensive hyperparameters optimization proves successful, the original formulation of \cite{howToActiveLearn} likely was sufficient and the image classification usecase just needs a high amount of fine-tuning.

\chapter{Future Work}\label{sec:futureWork}
\paragraph{Evaluation}
The framework in this case is considered to include the evaluation procedure for the trained agent. While the evaluation is clearly showing a lack of performance in the context of traditional active learning, it misses a comparison to any parameterized or end-to-end learned method. Such a comparison might give further insights on the state of the current system and how much it can be improved.
\paragraph{Environment}
Central part of the framework is of course the RL environment, which presents the following angles for improvements: 
(i) It is basically random for any state if it will receive a non-zero reward or not, since a reward is only issued when the budget is exhausted or a sub-game is finished and not depending on a special state/action pair (as it would be for a labyrinth environment).
(ii) The environment does not induce any form of labeling cost, which is a common theme in AL problems
(iii) The agent does not receive any negative reward for indefinitely rejecting every presented image sample
(iv) The current state representation does not include potentially important information like
\begin{itemize}
 \item The an embedded form of the presented images, i.e. by a pretrained feature extractor
 \item The number added images per class (and possibly metrics like the standard deviation of that distribution)
 \item Any representation of previous decisions or states
 \item The number of rejected samples since the last accepted one
\end{itemize}
\paragraph{Hyperparameters}
Due to time constraints, a number of technical details can be improved upon. Many hyperparameters of the training have not been optimized with a proper search technique. This includes the type of optimizer and parameters thereof, the architecture of the RL agents, the progression of the learning rate and greed parameter $\tau$ for the agent, as well as the momentum of the moving average for the internal F1-Score of the environment. \paragraph{Training Speed}
Furthermore, the overall speed of the training can be improved by fine tuning the fitting of the image classification model that takes place whenever an image is added to the labeled set. One possible solution would be to use a pretrained model and only apply transfer learning on the updated labeled set.
\paragraph{Number of Iterations}
Compared to other applications of RL, all experiments ran for a small number of interactions (8000-12000). An extensive experiment with an increased number of iterations might yield better results. \\
Furthermore, the training did not contain a ``warm-up'' phase where the agents collects experiences without actually training on them in order to populate the memory buffer.
\paragraph{Advanced Concepts}
Apart from technical implementations, multiple advanced concepts can be applied to the problem. \\
(i) An advanced memory replay can be employed by implementing the work of ``Prioritized Experience Replay'' by Schaul et. al. \cite{prioReplay}. \\
(ii) Following the ideas of curriculum learning \cite{currLearning}, the agent may be trained with problems of increasing difficulty by slowly increasing the budget and/or the length of the sub-games. \\
(iii) The advanced sampling strategy from \cite{rwAL4} can be used to quickly grow the labeled set and stabilize the CNN training, by automatically adding instances with high confidence.

\bibliographystyle{alpha}
\bibliography{thorben}

\newcommand{\etalchar}[1]{$^{#1}$}
\begin{thebibliography}{MKS{\etalchar{+}}15}

\bibitem[FLC17]{howToActiveLearn}
M.~Fang, Y.~Li, and T.~Cohn.
\newblock Learning how to active learn: A deep reinforcement learning approach.
\newblock {\em EMNLP 2017}, 2017.

\bibitem[FZL13]{newAlSurvey}
Yifan Fu, Xingquan Zhu, and Bin Li.
\newblock A survey on instance selection for active learning.
\newblock {\em Knowledge and information systems}, 35(2):249--283, 2013.

\bibitem[GBC16]{bookDeepLearning}
Ian Goodfellow, Yoshua Bengio, and Aaron Courville.
\newblock {\em Deep Learning}.
\newblock MIT Press, 2016.

\bibitem[GIG17]{rwAL5}
Yarin Gal, Riashat Islam, and Zoubin Ghahramani.
\newblock Deep bayesian active learning with image data.
\newblock {\em Proceedings of the 34th International Conference on Machine
  Learning-Volume 70}, 2017.

\bibitem[Goo]{googleLabeling}
Google.
\newblock Google cloud ai labeling service.

\bibitem[HJZL06]{rwAL3}
Steven~CH Hoi, Rong Jin, Jianke Zhu, and Michael~R Lyu.
\newblock Batch mode active learning and its application to medical image
  classification.
\newblock pages 417--424, 2006.

\bibitem[JPP09]{rwAL1}
Ajay~J Joshi, Fatih Porikli, and Nikolaos Papanikolopoulos.
\newblock Multi-class active learning for image classification.
\newblock pages 2372--2379, 2009.

\bibitem[KB06]{currLearning}
George Konidaris and Andrew Barto.
\newblock Autonomous shaping: Knowledge transfer in reinforcement learning.
\newblock {\em Proceedings of the 23rd international conference on Machine
  learning}, 2006.

\bibitem[LB95]{origConvNet}
Yann LeCun and Yoshua Bengio.
\newblock Convolutional networks for images, speech, and time series.
\newblock {\em The handbook of brain theory and neural networks}, 1995.

\bibitem[LMM09]{rwRL1}
Manuel Lopes, Francisco Melo, and Luis Montesano.
\newblock Active learning for reward estimation in inverse reinforcement
  learning.
\newblock {\em Joint European Conference on Machine Learning and Knowledge
  Discovery in Databases}, 2009.

\bibitem[LMP01]{CRF}
John Lafferty, Andrew McCallum, and Fernando~CN Pereira.
\newblock Conditional random fields: Probabilistic models for segmenting and
  labeling sequence data.
\newblock {\em Proceedings of the Eighteenth International Conference on
  Machine Learning (ICML)}, pages 282–--289, 2001.

\bibitem[MKS{\etalchar{+}}15]{mnih2015}
Volodymyr Mnih, Koray Kavukcuoglu, David Silver, Andrei~A Rusu, Joel Veness,
  Marc~G Bellemare, Alex Graves, Martin Riedmiller, Andreas~K Fidjeland, Georg
  Ostrovski, et~al.
\newblock Human-level control through deep reinforcement learning.
\newblock {\em nature}, 518(7540):529--533, 2015.

\bibitem[Ope]{openAiGym}
OpenAI.
\newblock Openai gym.

\bibitem[SB18]{bookRL}
R~Sutton and A~Barto.
\newblock {\em Reinforcement Learning: An Introduction}.
\newblock MIT Press, 2018.

\bibitem[Set09]{alSurvey}
Burr Settles.
\newblock Active learning literature survey.
\newblock 2009.

\bibitem[Set12]{alBook}
Burr Settles.
\newblock {\em Active Learning}, volume~6.
\newblock 2012.

\bibitem[Sha]{reluImg}
Daria Shamrai.
\newblock What is leaky relu?

\bibitem[SQAS15]{prioReplay}
Tom Schaul, John Quan, Ioannis Antonoglou, and David Silver.
\newblock Prioritized experience replay.
\newblock {\em arXiv preprint arXiv:1511.05952}, 2015.

\bibitem[TVC{\etalchar{+}}11]{rwAL2}
Devis Tuia, Michele Volpi, Loris Copa, Mikhail Kanevski, and Jordi Munoz-Mari.
\newblock A survey of active learning algorithms for supervised remote sensing
  image classification.
\newblock {\em IEEE Journal of Selected Topics in Signal Processing},
  5(3):606--617, 2011.

\bibitem[VHGS16]{ddqn}
Hado Van~Hasselt, Arthur Guez, and David Silver.
\newblock Deep reinforcement learning with double q-learning.
\newblock {\em Thirtieth AAAI conference on artificial intelligence}, 2016.

\bibitem[WZL{\etalchar{+}}]{rwAL4}
K.~Wang, D.~Zhang, Y.~Li, R.~Zhang, and L.~Lin.
\newblock Cost-effective active learning for deep image classification.
\newblock {\em IEEE Transactions on Circuits and Systems for Video Technology},
  27(12):2591--2600.

\end{thebibliography}

\addtocontents{toc}{\protect\setcounter{tocdepth}{1}}
\chapter{Appendix}

\section{Full Project}
The full project is available under\\
\href{https://github.com/ex0Therm1C/FinalExperiments}{https://github.com/ex0Therm1C/FinalExperiments}

\section{Experiment 1 Full Setup}

\subsection{Image Classifier}
\begin{tabular}{l | l | c | l }
Layer 1 & Layer 2 & Layer 3 & Layer 4 \\
\hline
Conv2D 64 & Conv2D 32 & Flatten & Dense 24 \\
Size: 3 & Size: 3 && \\
Stride: 3 & & &  \\
ReLU & ReLU  && ReLU
\end{tabular}

\subsection{Agent}
\begin{tabular}{l | c | l | c}
    Layer1 & Layer2 & Layer3 & Layer 4 \\
    \hline
    Dense 24 & BatchNorm & Dense 12 & BatchNorm \\
    LeakyReLU && LeakyReLU \\
    0.001 L2-Reg && 0.001 L2-Reg \\
    Init.: HE-uniform && Init.: HE-uniform \\
\end{tabular}\\[2mm]
\textbf{Policy:} Softmax Greedy \\
\textbf{Update Eq.} Van Hasselt \cite{ddqn} DDQN | $\gamma=0.9$\\
\textbf{Optimizer:} SGD \\

\subsection{Environment}
\paragraph{State Space} Vector $\in \mathbb{R}^{27}$ generated from a single image, containing (i) entropy and BvsSB score of the IC model's prediction, (ii) IC model layer-wise std-deviation, mean and norm, (iii) average F1-Score of the current IC model
\paragraph{Action Space} $a \in \{label, notLabel\}$

\subsection{General Parameters}
\begin{tabular}{c | c}
    \hline
    icModelMaxEpochs & 50\\
    earlyStoppingPatience & 1\\
    agentBatchSize & 64 \\
    targetNetworkUpdateRate (C) & 10 \\
    budget & 800 \\
    Reward Shaping & False \\
    maxInteractionPerGame & 1200 \\
    minTrainingInteractions & 12000 \\
    greedParameterRange & [1, 0.2]\\
    agentLearningRateRange & [0.001, 0.00001] \\
    exploration & 4000 \\
    conversion & 4000 \\
    memoryMaxLength & 1000
\end{tabular}

\vspace{10mm}
\section{Experiment 2 Full Setup}

\subsection{Image Classifier}
Unchanged from Experiment 1

\subsection{Agent}
\begin{tabular}{l | c | l | c}
    Layer1 & Layer2 & Layer3 & Layer 4 \\
    \hline
    Dense 48 & BatchNorm & Dense 24 & BatchNorm \\
    LeakyReLU && LeakyReLU \\
    0.001 L2-Reg && 0.001 L2-Reg \\
    Init.: HE-uniform && Init.: HE-uniform \\
\end{tabular}\\[2mm]
\textbf{Policy:} Softmax Greedy \\
\textbf{Update Eq.} Van Hasselt \cite{ddqn} DDQN | $\gamma=0.9$\\
\textbf{Optimizer:} SGD \\

\subsection{Environment}
\paragraph{State Space} Vector $\in \mathbb{R}^{35}$ generated from a sample of images, containing (i) entropy and BvsSB score of the IC model's prediction for each image, (ii) IC model's layer-wise std-deviation, mean and norm, (iii) average F1-Score of the current IC model
\paragraph{Action Space} $a \in [0, 1,  ... \hspace{1mm} , s + 1]$

\subsection{General Parameters}
\begin{tabular}{c | c}
    \hline
    icModelMaxEpochs & 50\\
    earlyStoppingPatience & 1\\
    agentBatchSize & 64 \\
    targetNetworkUpdateRate (C) & 10 \\
    budget & 800 \\
    Reward Shaping & False \\
    sample size & 5 \\
    maxInteractionPerGame & 1200 \\
    minTrainingInteractions & 12000 \\
    greedParameterRange & [1, 0.2]\\
    agentLearningRateRange & [0.001, 0.00001] \\
    exploration & 4000 \\
    conversion & 4000 \\
    memoryMaxLength & 1000
\end{tabular}

\vspace{10mm}
\section{Experiment 3 Full Setup}

\subsection{Image Classifier}
Unchanged from Experiment 1

\subsection{Agent}
Unchanged from Experiment 2

\subsection{Environment}
\paragraph{State Space} Vector $\in \mathbb{R}^{35}$ generated from a sample of images, containing (i) averaged entropy and BvsSB scores of the IC model's prediction for each image bundle, (ii) IC model's layer-wise std-deviation, mean and norm, (iii) average F1-Score of the current IC model
\paragraph{Action Space} $a \in [0, 1,  ... \hspace{1mm} , s + 1]$

\subsection{General Parameters}
\begin{tabular}{c | c}
    \hline
    icModelMaxEpochs & 50\\
    earlyStoppingPatience & 1\\
    agentBatchSize & 64 \\
    targetNetworkUpdateRate (C) & 10 \\
    budget & 800 \\
    Reward Shaping & False \\
    sample size & 5 \\
    Images to Bundle & 5 \\
    Sub-Game Length & 50 \\
    Reward Scaling & 40 \\
    maxInteractionPerGame & 1200 \\
    minTrainingInteractions & 8000 \\
    greedParameterRange & [1, 0.2]\\
    agentLearningRateRange & [0.001, 0.00001] \\
    exploration & 3000 \\
    conversion & 3000 \\
    memoryMaxLength & 1000
\end{tabular}

\addtocontents{toc}{\protect\setcounter{tocdepth}{2}}

\end{document}